\documentclass[journal]{IEEEtran}
\usepackage{graphicx}
\usepackage{amsmath, amssymb}
\usepackage{caption, subcaption}
\usepackage{float}
\usepackage{multirow}
\usepackage{hyperref}
\usepackage{soul,color}
\begin{document}
%%%%%%%%%%%%%%%%%%%%%%%%%%%%%%%%%%%%
\title{Hyperspectral Image Classification: Artifacts of Dimension Reduction on Hybrid CNN}
%%%%%%%%%%%%%%%%%%%%%%%%%%%%%%%%%%%%
\author{Muhammad Ahmad, Sidrah Shabbir, Rana Aamir Raza, Manuel Mazzara, Salvatore Distefano, Adil Mehmood Khan
\thanks{M. Ahmad is with the Department of Computer Science, National University of Computer and Emerging Sciences, Islamabad, Chiniot-Faisalabad Campus, Chiniot 35400, Pakistan and  Dipartimento di Matematica e Informatica---MIFT, University of Messina, Messina 98121, Italy. E-mail: mahmad00@gmail.com} 
\thanks{S. Shabbir is with the Department of Computer Engineering, Khwaja Fareed University of Engineering and Information~Technology, Rahim Yar Khan, 64200, Pakistan; Email: sidrah.shabbir@gmail.com}
\thanks{R. A. Raza is with the Department of Computer Science, Bahauddin Zakariya University, Multan 66000, Pakistan; E-mail: aamir@bzu.edu.pk}
\thanks{M. Mazzara is with Innopolis University, Innopolis 420500, Russia; E-mail: m.mazzara@innopolis.ru}
\thanks{S. Distefano is with  Dipartimento di Matematica e Informatica---MIFT, University of Messina, Messina 98121, Italy; E-mail: sdistefano@unime.it}
\thanks{A. M. Khan is with Innopolis University, Innopolis 420500, Russia; E-mail: a.khan@innopolis.ru}
}
%%%%%%%%%%%%%%%%%%%%%%%%%%%%%%%%%%%%
\markboth{Preprint Submitted to arXiv, January~2021}
{M.Ahmad \MakeLowercase{\textit{et al.}}: Artifacts of DR on CNN for HSIC}
%%%%%%%%%%%%%%%%%%%%%%%%%%%%%%%%%%%%
\maketitle
%%%%%%%%%%%%%%%%%%%%%%%%%%%%%%%%%%%%%%%%%%
\begin{abstract}
Convolutional Neural Networks (CNN) has been extensively studied for Hyperspectral Image Classification (HSIC) more specifically, 2D and 3D CNN models have proved highly efficient in exploiting the spatial and spectral information of Hyperspectral Images. However, 2D CNN only considers the spatial information and ignores the spectral information whereas 3D CNN jointly exploits spatial-spectral information at a high computational cost. Therefore, this work proposed a lightweight CNN (3D followed by 2D-CNN) model which significantly reduces the computational cost by distributing spatial-spectral feature extraction across a lighter model alongside a preprocessing that has been carried out to improve the classification results. Five benchmark Hyperspectral datasets (i.e., SalinasA, Salinas, Indian Pines, Pavia University, Pavia Center, and Botswana) are used for experimental evaluation. The experimental results show that the proposed pipeline outperformed in terms of generalization performance, statistical significance, and computational complexity, as compared to the state-of-the-art 2D/3D CNN models except commonly used computationally expensive design choices.
\end{abstract}
%%%%%%%%%%%%%%%%%%%%%%%%%%%%%%%%%%%%%%%%%%
%%%%%%%%%%%%%%%%%%%%%%%%%%%%%%%%%%%%
\IEEEpeerreviewmaketitle
%%%%%%%%%%%%%%%%%%%%%%%%%%%%%%%%%%%%%%%%%%
\section{Introduction}

Hyperspectral Imaging (HSI) records the reflectance values of a target object across a wide range of electromagnetic spectrum instead of just visible range (red, green, blue (RGB) channels) \cite{ahmad2020spatial}. Light interacting with each pixel is fractionated into hundreds of contiguous spectral bands in order to provide sufficient information about the target. The high spectral resolution of HSI allows us to distinguish various objects based on spectral signatures. The conventional classifiers used for hyperspectral image classification (HSIC) are SVM \cite{wang2020multiple}, k-Nearest Neighbors (KNN) \cite{ma2020graph, ahmad2020multiclass}, Maximum Likelihood \cite{alcolea2020inference}, Logistic Regression \cite{ahmad2019spatial}, Random Forests (RF) \cite{xia2013hyperspectral}, Multinomial Logistic Regression (MLR), Extreme Learning Machine (ELM) \cite{ahmad2020fuzziness}, and 3D Convolution Neural Networks (CNN) \cite{ahmad2020fast}. These traditional classification methods classify the HSI based on spectral signatures only, and therefore, do not outperform owing to redundancy present in spectral information and high correlation among spectral bands. Moreover, these methods fail to maintain the spatial variability of HSI that also hinders the HSIC performance. 

HSIC performance can be enhanced by considering two main aspects: dimensionality reduction and utilization of spatial information contained in HSI. Dimensionality reduction is an important preprocessing step in HSIC to reduce the spectral redundancy of HSI that subsequently results in less processing time and enhanced classification accuracy. Dimensionality reduction methods transform the high-dimensional data into a low-dimensional space whilst preserving the potential spectral information \cite{AHMAD2019370}. 

Spatial information improves the discriminative power of the classifier by considering the neighboring pixels’ information. Generally, spectral-spatial classification approaches can be categorized into two groups. The first type excavates for both spectral and spatial features individually. Spatial information is educed in advanced using various methods like morphological operations \cite{benediktsson2005classification}, attribute profiles \cite{dalla2010classification} and entropy \cite{tuia2014automatic, AHMAD2021166267} etc., and then spliced together with spectral features for pixels-wise classification. The other type coalesces spectral and spatial information to acquire joint features like Gabor filter and wavelets \cite{shen2011three, qian2012hyperspectral} are constructed at various scales to simultaneously extract spectral-spatial features for HSIC.

Traditional feature extraction approaches are based on handcrafted features and usually extract shallow HSI features. Moreover, such approaches rely on a high level of domain knowledge for feature designing \cite{wang2020novel}. To overcome the limitations of traditional feature extraction techniques, Deep learning (DL) has been widely used to automatically learn the low to a high-level representation of HSI in a hierarchical manner \cite{Yang20, Roy20}. DL based HSIC frameworks present enhanced generalization capabilities and improved predictive performance \cite{li2019deep}. 

Recently, Convolutional Neural Network (CNN) has proven to be a powerful feature extraction tool that can learn effective features of HSI through a network of hidden layers \cite{shabbir2021hyperspectral}. CNN based HSIC architectures have attracted prevalent attention due to substantial performance gain. Generally, the 2D CNN are utilized for efficient spatial feature exploitation and to extract both spectral and spatial features of HSI, many variants of 3D CNN have been proposed \cite{ wang2019classification, paoletti2018new, li2017spectral, he2017multi}. However, 3D CNN is a computationally complex model and 2D CNN alone cannot efficiently extract discriminating spectral features. 

To overcome these challenges, a hybrid 3D/2D CNN model that splice together 3D CNN components with 2D CNN components/layers is proposed. The aim is to synergize the efficacies of 3D CNN and 2D CNN to obtain important discriminating spectral-spatial features of HSI for HSIC. Moreover, we incorporated the dimensionality reduction as a preprocessing step and investigated the impact of various state-of-the-art dimensionality reduction approaches on the performance of the hybrid 3D/2D CNN model. Also, we evaluated the impact of different input window sizes on the performance of the proposed framework. A comparative study is also carried out using several state-of-the-art CNN based HSIC frameworks proposed in recent literature. Experimental results on five benchmark datasets revealed that our proposed hybrid 3D/2D CNN model outperformed the compared HSIC frameworks.

The rest of the paper is structured as follows: Section \ref{Sec:2} presents the proposed pipeline including the details of each component implemented in this paper. Section \ref{Sec:3} describes the important experimental settings and evaluation metrics. Section \ref{Sec:4} contains information regarding the experimental datasets and results. Section \ref{Sec:5}presents the important discussion on obtained results. Section \ref{Sec:6} describes the experimental comparisons with state-of-the-art methods. Finally, Section \ref{Sec:7} concludes the paper with possible future research directions.

%%%%%%%%%%%%%%%%%%%%%%%%%%%%%%%%%%
\section{Proposed Methodology}
\label{Sec:2}

Let us assume that the HSI cube can be represented as $\textbf{X} = [x_1, x_2, x_3, \dots, x_S]^T \in \mathcal{R}^{S \times (C \times D)}$, where $S$ denotes total number of spectral bands and $(C \times D)$ are the samples per band belonging to $Y$ classes and $x_i = [x_{1,i},~x_{2,i},~x_{3,i},~\dots,x_{S,i}]^T$ is the $i^{th}$ sample in the HSI cube. Suppose $(x_i, y_i) \in (\mathcal{R}^{S \times (C \times D)} , \mathcal{R}^Y)$, where $y_i$ is the class label of the $i^{th}$ sample. However, due to the spectral mixing effect which induces high intra-class variability and high inter-class similarity in HSI, it becomes difficult to classify various materials based on their spectral signatures. To combat the aforesaid issue, we used dimensionality reduction as a preprocessing step to eliminate the spectral redundancy of HSI which reduces the number of spectral bands $(S \to B$, where $B \ll S)$ while keeping the spatial dimensions unimpaired. Subsequently, this also results in a reduced computational overhead owing to a lower-dimensional feature subspace. We evaluated our proposed model with several dimensionality reduction approaches defined in section \ref{Sec:BS}.

%%%%%%%%%%%%%%%%%%%%%%%%%%%%%%%%%%
\subsection{Dimensionality Reduction Methods}
\label{Sec:BS}

Dimensionality reduction is an essential preprocessing step in HSI analysis that can be categories as either feature extraction or selection. Feature extraction transforms the high dimensional HSI data into a low dimensional subspace by extracting suitable feature representation that can give classification performance comparable to the model trained on the original set of spectral bands while band selection approaches extract a subset of discriminative bands to reduce spectral redundancy. In this work, we evaluated the effectiveness of the following dimensionality reduction methods for our proposed hybrid 3D/2D CNN model.

%%%%%%%%%%%%%%%%%%%%%%%%%%%%%%%%%%
\subsubsection{Principle Component Analysis (PCA)}

Principal Component Analysis (PCA) is a feature extraction technique based on the orthogonal transformation that computes linearly uncorrelated variables, known as principal components (PCs), from possibly correlated data. The first PC is the projection on the direction of highest variance and it gradually decreases as we move towards the last PC. The transformation of the original image to PCs is the Eigen decomposition of the covariance matrix of mean-centered HSI data. Eigen decomposition of covariance matrix i.e. finding the eigenvalues along with their corresponding eigenvectors is as follows:

\begin{equation*}
   E = ADA^{T}
\end{equation*}
where $A = {a_1, a_2, a_3 \dots, a_S}$ is a transformation matrix and $D = diag{\lambda_1, \lambda_2, \lambda_3, \dots, \lambda_S}$ is a diagonal matrix of eigenvalues of covariance matrix. The linear HSI transformation is defined as:

\begin{equation*}
    H = AX
\end{equation*}

%%%%%%%%%%%%%%%%%%%%%%%%%%%%%%%%%%
\subsubsection{Incremental PCA (iPCA)}

Generally, PCA is performed in batch mode i.e. all the training data is simultaneously available to compute the projection matrix. In order to find the updated PCs after incorporating the new data into the existing training set, PCA needs to be retrained with complete training data. To combat this limitation, an incremental PCA (iPCA) approach is used that can be categorized as either covariance-based iPCA or covariance free iPCA method. Covariance based methods are further divided into two approaches. In the first approach, the covariance matrix is computed using existing training data and then the matrix is updated whenever new data samples are added. In the second approach, a reduced covariance matrix is computed using previous PCs and the new training data. Covariance free iPCA methods update the PCs without computing the covariance methods, however, such methods usually face convergence problems in case of high dimensional data \cite{weng2003candid}.

%%%%%%%%%%%%%%%%%%%%%%%%%%%%%%%%%%
\subsubsection{Sparse PCA (SPCA)}

The conventional PCA has a limitation that the PCs are the linear combinations of all input features/predictors or in other words, all the components are nonzero and direct interpretation becomes difficult. Therefore, to improve interpretability, it is desirable to use sparsity promoting regularizers. In this regard, sparse PCA (SPCA) has emerged as an effective technique that finds the linear combinations of a few inputs features i.e. only a few active (nonzero) coefficients. SPCA works well in the scenarios where input features are redundant, that is, they do not contribute to identifying the underlying rational model structure.

%%%%%%%%%%%%%%%%%%%%%%%%%%%%%%%%%%
\subsubsection{Singular Value Decomposition (SVD)}

Singular Value Decomposition (SVD) is a mathematical technique that decomposes a matrix into three different matrices. It is knowns as truncated SVD when used for dimensionality reduction. This matrix decomposition is represented as:

\begin{equation*}
X = PSQ^{T} 
\end{equation*} 
where $P$ and $Q$ are orthogonal matrices of left and right singular vectors and $S$ is a diagonal matrix having singular values as its diagonal entries. An SVD reduced $X$ is obtained by taking into account the contribution of only the first $k$ eigenimages, computed as follows:

\begin{equation*}
X_{SVD} = \sum_{i=1}^{k} P_{i}S_{i}Q_{i}^{T} 
\end{equation*} 

%%%%%%%%%%%%%%%%%%%%%%%%%%%%%%%%%%
\subsubsection{Independent Component Analysis (ICA)}

Independent Component Analysis (ICA) is one of the popular approaches among other dimensionality reduction methods, that extracts statistically independent components (ICs) through a linear or non-linear transformation that minimizes the mutual information between ICs or maximizes the likelihood or non-Gaussianity of ICs. It transforms the HSI into a lower-dimensional feature space by comparing the average absolute weight coefficients for each spectral band of HSI and retain only those independent bands which contain maximum information \cite{du2003band}. Given an n-dimensional data $X$, the main task of ICA is to find the linear transformation $W$ such that:

\begin{equation*}
    H = WX
\end{equation*}
where $H$ has statistically independent components. 

%%%%%%%%%%%%%%%%%%%%%%%%%%%%%%%%%%
% \subsubsection{Gaussian Random Projection (GRP)}

% Random projection projects the high dimensional data into lower-dimensional subspace by maintaining the distance between the data samples. Gaussian Random projection (GRP) is one of the commonly utilized and computationally efficient random projection approaches. For a n-dimensional data $X^{m \times n}$, the GRP maps $X$ into a lower-dimensional data $H^{m \times k}$ with dimensions equal to  $k$ (where $k \ll n$) by using a Gaussian projection matrix $R^{n \times k}$. The Gaussian projection matrix is created from Gaussian distribution, which satisfies the properties of orthogonality and normality. The projection of the data can be represented as:

% \begin{equation*}
%     H^{m \times k}  = X^{m \times n} R^{n \times k}
% \end{equation*}

%%%%%%%%%%%%%%%%%%%%%%%%%%%%%%%%%%
\subsection{Hybrid 3D/2D CNN}

In order to pass the HSI data cube to our Hybrid CNN model, it is divided into multiple small overlapping 3D patches, and the class labels of these patches are decided based on the label of central pixel. The $3D$ neighboring patches $P \in \mathcal{R}^{(W \times W) \times B}$ are formed that are centered at spatial position $(a, b)$, covering the $W \times W$ windows. The total number of 3D spatial patches created from $X$ is given by $(M - W + 1) \times (N - W + 1)$. These $3D$ patches centered at location $(a, b)$ represented by $P_{(a, b)}$ covers the width from $\frac{a - (W - 1)}{2}$ to $\frac{a + (W - 1)}{2}$ and height from $\frac{b – (W - 1)}{2}$ to $\frac{b + (W - 1)}{2}$ and all $B$ spectral bands obtained after dimensionality reduction method.

In the 2D CNN, input data is convolved with the 2D kernel function that computes the sum of the dot product between the input and the 2D kernel function. The kernel is stridden over the input in order to cover the whole spatial dimension. Then these convolved features are processed through an activation function that helps to learn non-linear features of data by introducing non-linearity in the model. In case of 2D convolution, the activation value of $j^{th}$ feature map at spatial location $(x, y)$ in the $i^{th}$ layer, denoted by $v^{x, y}_{i, j}$, can be formulated as follows:

\begin{equation*}
v^{x, y}_{i, j} = \mathcal{F} ( b_{i, j} \sum_{\tau=1}^{d_{l-1}} \sum_{\rho = -\gamma}^{\gamma} \sum_{\sigma = -\delta}^{\delta} w_{i, j, \tau}^{\sigma, \rho} \times v_{i-1, \tau}^{x+\sigma, y+\rho} )
\label{Eq1}
\end{equation*}
where $\mathcal{F}$ is the activation function, $d_{l-1}$ is the number of feature map at $(l-1)^{th}$ layer and the depth of kernel $w_{i, j}$ for $j^{th}$ feature map at $i^{th}$ layer, $b_{i, j}$ denotes the bias parameter for $j^{th}$ feature map at $i^{th}$ layer, $2\gamma + 1$ and $2\sigma + 1$ be the width and height of the kernel.

The 3D convolutional process first computes the sum of the dot product between input patches and 3D kernel function i.e. the 3D input patches are convolved with 3D kernel function \cite{Yang20, Roy20}. Later these feature maps are passed through an activation function to induce non-linearity. Our proposed model generates the features maps of the 3D convolutional layer by using 3D kernel function over $B$ spectral bands, extracted after dimensionality reduction, in the input layer. In the 3D convolutional process of the proposed model, the activation value at spatial location $(x, y, z)$ at the $i^{th}$ layer and $j^{th}$ feature map can be formulated as:

\begin{equation*}
v^{x, y}_{i, j} = \mathcal{F} ( b_{i, j} \sum_{\tau=1}^{d_{l-1}} \sum_{\lambda = -v}^{v} \sum_{\rho = -\gamma}^{\gamma} \sum_{\sigma = -\delta}^{\delta} w_{i, j, \tau}^{\sigma, \rho, \lambda} \times v_{i-1, \tau}^{x+\sigma, y+\rho, z+\lambda} )
\label{Eq2}
\end{equation*}
where all the parameters are the same as defined in Equation \ref{Eq1} except $2v + 1$ which is the depth of 3D kernel along a spectral dimension. In the proposed framework, the details of 3D convolutional kernels are as follows: $3D\_conv\_layer\_1 = 8 \times 3 \times 3 \times 7 \times 1$ i.e. $K_{1}^{1} = 3, K_{2}^{1} = 3$ and $K_{3}^{1} = 7$. $3D\_conv\_layer\_2 = 16 \times 3 \times 3 \times 5 \times 8$  i.e. $K_{1}^{2} = 3, K_{2}^{2} = 3$ and $K_{3}^{2} = 5$. $3D\_conv\_layer\_3 = 32 \times 3 \times 3 \times 3 \times 16$ i.e. $K_{1}^{3} = 3, K_{2}^{3}= 3$ and $K_{3}^{3} = 3$. The details of $2D$ convolutional kernel are:  $2D\_conv\_layer\_1 = 64 \times 3 \times 3 \times 96$ i.e. $K_{1}^{4} = 3$ and $K_{2}^{4} = 3$. Three $3D$ convolutional layers are employed to increase the number of spectral-spatial feature maps and one $2D$ convolutional layer is used to discriminate the spatial features within different spectral bands while preserving the spectral information. 

\begin{figure*}[!hbt]
    \centering
    \includegraphics[width=0.99\textwidth]{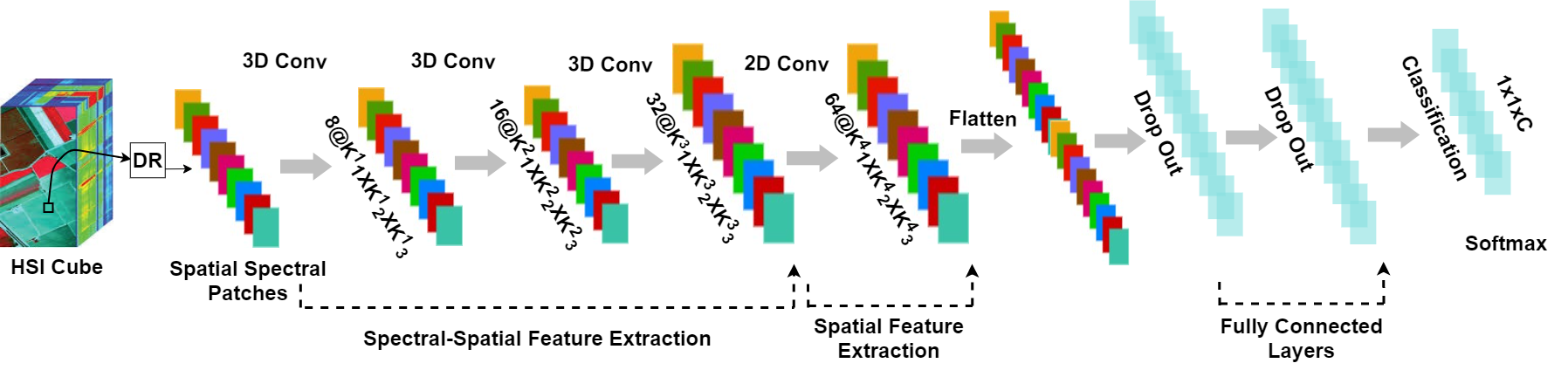}
    \caption{Hybrid 3D/2D CNN framework for HSIC}
    \label{Fig.model}
\end{figure*}

Further details regarding the Hybrid 3D/2D CNN architecture in terms of types of layers, dimensions of output feature maps, and a number of trainable parameters are given in Table \ref{Tab.1A} and layer-wise hierarchy is shown in Figure \ref{Fig.model}. The total number of tune-able weight parameters of our proposed model is $127,104$ for the Salinas Full Scene dataset. Initially, the weights are randomized and then optimized using backpropagation with the Adam optimizer by using the soft max loss function. The network is trained for $50$ epochs using a mini-batch size of $256$ and without any batch normalization and data augmentation.

%%%%%%%%%%%%%%%%%%%%%%%%%%%%%%%%%%%%
\begin{table}[!hbt]
    \caption{Layer wise detailed Summary Hybrid 3D/2D CNN architecture on Salinas Full Scene Dataset with Window Size $9 \times 9$ and $15$ bands.}
    \centering
    \begin{tabular}{c|c|c}  \hline
    \textbf{Layer} & \textbf{Output Shape} & \textbf{$\#$ of Parameters} \\ \hline
    Input Layer & (9, 9, 15, 1) & 0 \\
    Conv3D\_1 (Conv3D) & (7, 7, 9, 8) & 512 \\
    Conv3D\_2 (Conv3D) & (5, 5, 5, 16) & 5776  \\
    Conv3D\_3 (Conv3D) & (3, 3, 3, 32) & 13856 \\
    Reshape (Reshape) & (3, 3, 96) & 0 \\
    Conv2D\_1 (Conv2D) & (1, 1, 64) & 55360 \\
    Flatten\_1 (Flatten) & (64)& 0 \\
    Dense\_1 (Dense) & (256) & 16640 \\
    Dropout\_1 (Dropout) & (256) & 0 \\
    Dense\_2 (Dense) & (128) & 32896 \\
    Dropout\_2 (Dropout) & (128) & 0 \\ 
    Dense\_3 (Dense) & (16) & 2064\\ \hline
    \multicolumn{3}{c}{In total, \textbf{127,104} trainable parameters are required}  \\ \hline
    \end{tabular}
    \label{Tab.1A}
\end{table}

%%%%%%%%%%%%%%%%%%%%%%%%%%%%%%%%%%%%%
%%%%%%%%%%%%%%%%%%%%%%%%%%%%%%%%%%

%%%%%%%%%%%%%%%%%%%%%%%%%%%%%%%%%%
\section{Experimental Settings}
\label{Sec:3}

The experimental datasets used in this manuscript are used for a variety of unmixing and classification tasks in which the number of bands, the number of samples, and the number of classes are different. Experimental validation is being carried out in the $K$-fold cross-validation process a base case to split the train/validation/test samples in which the value of $K$ is set to $5$.

To validate the experimental results and our proposed pipeline, we conduct several statistical tests such as Precision ($P_r)$, Recall ($R_c$), F1 Score, Overall (OA -- is computed as the number of correctly classified examples out of the total test examples), Average (AA -- represents the average class-wise classification performance), and Kappa ($\kappa$ -- is known as a statistical metric that considered the mutual information regarding a strong agreement among classification and ground-truth maps) accuracies with $95\%$ confidence interval to validate the claims made in this manuscript. The said metrics are computed using the following mathematical expressions:

\begin{equation*}
    P_r = \frac{1}{Y} \sum_{i = 1}^Y \frac{TP_i}{TP_i + FP_i}
\end{equation*}

\begin{equation*}
    R_c = \frac{1}{Y} \sum_{i = 1}^Y \frac{TP_i}{TP_i + FN_i}
\end{equation*}

\begin{equation*}
    F1 = \frac{2 \times (R_c \times P_r) }{(R_c + P_r)}
\end{equation*}

\begin{equation*}
    OA = \frac{1}{K} \sum_{i = 1}^K TP_i
\end{equation*}

\begin{equation*}
    \kappa = \frac{P_o - P_e}{1 - P_e}
\end{equation*}
where 
\begin{equation*}
    P_e = P^+ + P^-
\end{equation*}

\begin{equation*}
P^+ = \frac{TP + FN}{TP + FN + FP + TN} \times \frac{TP + FN}{TP + FN + FP + TN}
\end{equation*}

\begin{equation*}
P^- = \frac{FN + TN}{TP + FN + FP + TN} \times \frac{FP + TN}{TP + FN + FP + TN}
\end{equation*}

\begin{equation*}
    P_o = \frac{TP + TN}{TP + FN + FP + TN}
\end{equation*}
where $FP$ and $TP$ are false and true positive, $FN$ and $TN$ are false and true negative computed from the confusion matrix.

All the experiments are performed on Colab \cite{carneiro2018}. Colab provides an option to execute the codes on python $3$ notebook with GPU with $25$ GB RAM and $358.27$ GB of could storage for computations. In all the experiments, each dataset is initially divided into a $50/50\%$ ratio for the Training and Test set and then the training set is further split into a $50/50\%$ ratio for Training and Validation samples.

In all the experiments, learning rate is set to $0.001$ and activation function used for all the layers is $relu$ except the last layer where $softmax$ is applied. Spatial dimensions of 3D input patches for all datasets are set as as $9 \times 9 \times 15$, $11 \times 11 \times 15$, $9 \times 9 \times 18$, $11 \times 11 \times 18$, $9 \times 9 \times 21$, $11 \times 11 \times 21$, $9 \times 9 \times 24$, $11 \times 11 \times 24$, and $9 \times 9 \times 27$, $11 \times 11 \times 27$, where $15, 18, 21, 24$ and $27$ are the number of most informative bands extracted by PCA, iPCA, SPCA, ICA, SVD, and GRP.

%%%%%%%%%%%%%%%%%%%%%%%%%%%%%%%%%%%%
\section{Experimental Results}
\label{Sec:4}

The effectiveness of our proposed hybrid 3D/2D CNN is confirmed on five benchmark HSI datasets available publicly and acquired by two different sensors, i.e., Reflective Optics System Imaging Spectrometer (ROSIS) and Airborne Visible/Infrared Imaging Spectrometer (AVIRIS). These five datasets are, Salinas-A (SA), Salinas full scene (SFS), Indian Pines (IP), and Pavia University (PU).

%The effectiveness of our proposed hybrid 3D/2D CNN is confirmed on five benchmark HSI datasets available publicly and acquired by three different sensors, i.e., Reflective Optics System Imaging Spectrometer (ROSIS), Airborne Visible/Infrared Imaging Spectrometer (AVIRIS) and NASA EO-1 satellite Hyperion sensor. These five datasets are, Salinas-A (SA), Salinas full scene (SFS), Indian Pines (IP) and Pavia University (PU). %, and Botswana (BS).

%%%%%%%%%%%%%%%%%%%%%%%%%%%%%%%%%%
\subsection{Salinas-A Dataset (SA)}

\begin{figure}[!hbt]
    \centering
%     \begin{subfigure}{0.085\textwidth}
% 		\includegraphics[width=0.99\textwidth]{images/SLA_ground_truth.png}
% 		\centering
% 		\caption{SA} 
% 		\label{Fig.SA_gt}
% 	\end{subfigure}
    \begin{subfigure}{0.085\textwidth}
		\includegraphics[width=0.99\textwidth]{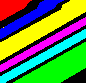}
		\centering
		\caption{$15,9$}
		\label{Fig.1A}
	\end{subfigure}
	\begin{subfigure}{0.085\textwidth}
		\includegraphics[width=0.99\textwidth]{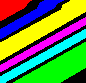}
		\centering
		\caption{$15,11$} 
		\label{Fig.1B}
	\end{subfigure}
	\begin{subfigure}{0.085\textwidth}
		\includegraphics[width=0.99\textwidth]{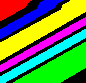}
		\centering
		\caption{$18,9$}
		\label{Fig.1C}
	\end{subfigure}
	\begin{subfigure}{0.085\textwidth}
		\includegraphics[width=0.99\textwidth]{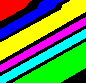}
		\centering
		\caption{$18,11$}
		\label{Fig.1D}
	\end{subfigure}
	\begin{subfigure}{0.085\textwidth}
		\includegraphics[width=0.99\textwidth]{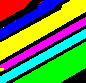}
		\centering
		\caption{$21,9$}
		\label{Fig.1E}
	\end{subfigure}
	\begin{subfigure}{0.085\textwidth}
		\includegraphics[width=0.99\textwidth]{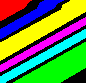}
		\centering
		\caption{$21,11$} 
		\label{Fig.1F}
	\end{subfigure}
	\begin{subfigure}{0.085\textwidth}
		\includegraphics[width=0.99\textwidth]{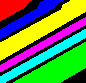}
		\centering
		\caption{$24,9$}
		\label{Fig.1G}
	\end{subfigure}
	\begin{subfigure}{0.085\textwidth}
		\includegraphics[width=0.99\textwidth]{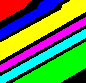}
		\centering
		\caption{$24,11$}
		\label{Fig.1H}
	\end{subfigure}
	\begin{subfigure}{0.085\textwidth}
		\includegraphics[width=0.99\textwidth]{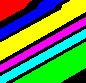}
		\centering
		\caption{$27,9$}
		\label{Fig.1I}
	\end{subfigure}
	\begin{subfigure}{0.085\textwidth}
		\includegraphics[width=0.99\textwidth]{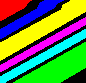}
		\centering
		\caption{$27,11$}
		\label{Fig.1J}
	\end{subfigure}
\caption{Classification results for \textbf{Salinas-A} for different number of bands ($15, 18, 21, 24, 27$) selected using PCA, with different number of patch sizes ($9\times9$ and $11\times11$).}
\label{Fig.1}
\end{figure}

\begin{figure}[!hbt]
	\begin{subfigure}{0.22\textwidth}
		\includegraphics[width=0.99\textwidth]{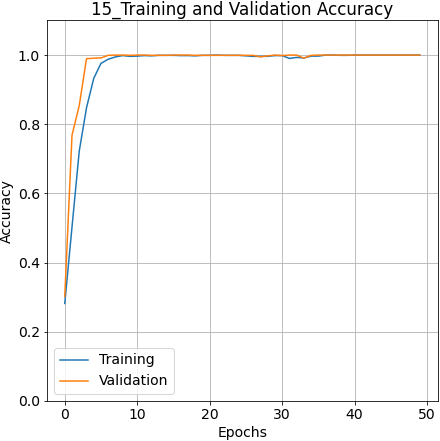}
		\centering
		\caption{Accuracy, $9\times9 \times 15$} 
		\label{Fig.SA_LCA}
	\end{subfigure}
	\begin{subfigure}{0.22\textwidth}
		\includegraphics[width=0.99\textwidth]{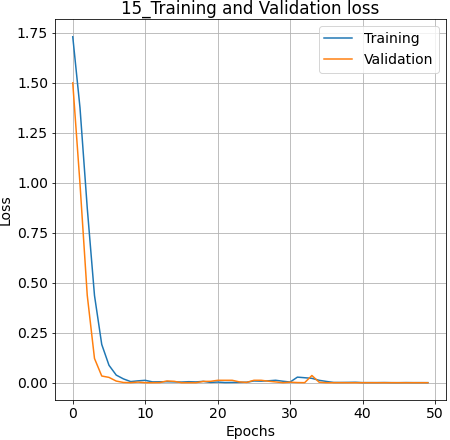}
		\centering
		\caption{Loss, $9\times9 \times 15$}
		\label{Fig.SA_LCB}
	\end{subfigure}
	
	\begin{subfigure}{0.22\textwidth}
		\includegraphics[width=0.99\textwidth]{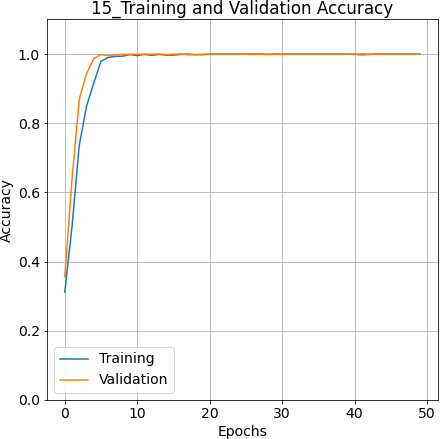}
		\centering
		\caption{Accuracy, $11\times11 \times 15$} 
		\label{Fig.SA_LCC}
	\end{subfigure}
	\begin{subfigure}{0.22\textwidth}
		\includegraphics[width=0.99\textwidth]{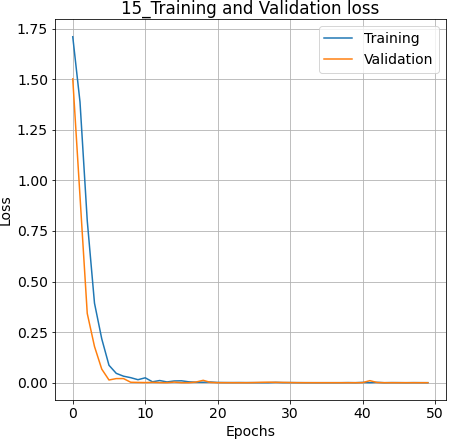}
		\centering
		\caption{Loss, $11\times11 \times 15$} 
		\label{Fig.SA_LCD}
	\end{subfigure}
\caption{Accuracy and Loss for  Training  and  Validation  sets on  \textbf{Salinas-A  Dataset} for $50$ number of epochs with two different spatial dimensions ($9\times 9$ and $11\times 11$) and $15$ number of bands.}
\label{Fig.SA_LC}
\end{figure}

Salinas-A (SA) is a sub-set of Salinas full scene (SFS) dataset consisting of $86\times83$ samples per band with a total of $224$ and have six classes as shown in Table \ref{tab:SA_stat}. In the full Salinas scene, SA is located at $158-240$ and $591-676$. A detailed experimental results on Salinas-A dataset is shown in Table \ref{tab:SA_acc} and Figure \ref{Fig.1}. Moreover, the statistical significance is shown in Table \ref{tab:SA_stat}. The convergence loss and accuracy of our proposed Hybrid $3D/2D$ CNN for $50$ epochs with two different patch sizes are illustrated in Figure \ref{Fig.SA_LC}. From these accuracy and loss curves, one can deduce that our proposed model is converged almost around $7$ epoch for both $9\times 9$ and $11\times 11$ window sizes.

%%%%%%%%%%%%%%%%%%%%%%%%%%%%%%%%%%%%%%%%%%%%
\begin{table}[!hbt]
\centering
    \caption{Kappa, Overall and Average accuracy for \textbf{Salinas-A dataset} with different number of bands (i.e., $15$, $18$, $21$, $24$, $27$, respectively) and different number of patch sizes (i.e., $9\times9$ and $11\times11$, respectively).}
    
    \resizebox{\columnwidth}{!}{
    \begin{tabular}{c|cc|cc|cc|cc|cc} \hline
    
    \multirow{2}{*}{\textbf{Method}} &  \multicolumn{2}{c|}{\textbf{15   Bands}} & \multicolumn{2}{c|}{\textbf{18   Bands}} & \multicolumn{2}{c|}{\textbf{21   Bands}} & \multicolumn{2}{c|}{\textbf{24   Bands}} & \multicolumn{2}{c}{\textbf{27   Bands}} \\ \cline{2-11} 
    
     & \textbf{9 $\times$ 9} & \textbf{11 $\times$ 11} & \textbf{9 $\times$ 9} & \textbf{11 $\times$ 11} & \textbf{9 $\times$ 9} & \textbf{11 $\times$ 11} & \textbf{9 $\times$ 9} & \textbf{11 $\times$ 11} & \textbf{9 $\times$ 9} & \textbf{11 $\times$ 11} \\ \hline
    
    \multirow{3}{*}{\textbf{PCA}} & 99.95 & 100.00 & 100.00 & 100.00 & 99.53  & 100.00 & 100.00 & 100.00 & 100.00 & 100.00 \\ \cline{2-11} 
 
    & 99.96 & 100.00 & 100.00 & 100.00 & 99.63  & 100.00 & 100.00 & 100.00 & 100.00 & 100.00 \\ \cline{2-11} 
    & 99.95 & 100.00 & 100.00 & 100.00 & 99.49  & 100.00 & 100.00 & 100.00 & 100.00 & 100.00 \\ \hline

    \multirow{3}{*}{\textbf{iPCA}} & 90.76 & 99.81  & 100.00 & 90.76  & 99.95  & 100.00 & 99.95  & 99.81  & 99.95  & 99.95  \\ \cline{2-11} 
    & 92.67 & 99.85  & 100.00 & 92.67  & 99.96  & 100.00 & 99.96  & 99.85  & 99.96  & 99.96  \\ \cline{2-11} 
    & 83.33 & 99.75  & 100.00 & 83.33  & 99.98  & 100.00 & 99.91  & 99.70  & 99.95  & 99.95  \\ \hline
    
    \multirow{3}{*}{\textbf{SPCA}} & 85.29 & 99.06  & 99.77  & 99.91  & 90.76  & 99.91  & 99.86  & 99.95  & 99.91  & 100.00 \\ \cline{2-11} 
    & 88.48 & 99.25  & 99.81  & 99.93  & 92.67  & 99.93  & 99.89  & 99.96  & 99.93  & 100.00 \\ \cline{2-11} 
    & 83.33 & 99.12  & 99.64  & 99.95  & 83.33  & 99.89  & 99.93  & 99.95  & 99.95  & 100.00 \\ \hline

    \multirow{3}{*}{\textbf{ICA}} & 99.86 & 99.95  & 100.00 & 99.95  & 99.91  & 100.00 & 100.00 & 100.00 & 100.00 & 100.00 \\ \cline{2-11} 
    & 99.89 & 99.96  & 100.00 & 99.96  & 99.93  & 100.00 & 100.00 & 100.00 & 100.00 & 100.00 \\ \cline{2-11} 
    & 99.84 & 99.95  & 100.00 & 99.95  & 99.83  & 100.00 & 100.00 & 100.00 & 100.00 & 100.00 \\ \hline

    \multirow{3}{*}{\textbf{SVD}}  & 99.67 & 0.00   & 0.00   & 74.69  & 100.00 & 75.13  & 99.81  & 84.81  & 75.35  & 49.97  \\ \cline{2-11} 
    & 99.74 & 28.50  & 28.50  & 79.99  & 100.00 & 80.67  & 99.85  & 88.03  & 80.93  & 59.99  \\ \cline{2-11} 
    & 99.65 & 16.67  & 16.67  & 66.56  & 100.00 & 66.28  & 99.81  & 82.47  & 66.52  & 66.67  \\ \hline
    
    %\multirow{3}{*}{\textbf{GRP}} & 0.00  & 0.00   & 0.00   & 0.00   & 0.00   & 0.00   & 0.00   & 0.00   & 0.00   & 0.00   \\ \cline{2-11} 
    %& 28.50 & 28.50  & 28.50  & 25.09  & 28.50  & 28.50  & 28.50  & 28.50  & 28.50  & 25.09  \\ \cline{2-11} 
    %& 16.67 & 16.67  & 16.67  & 16.67  & 16.67  & 16.67  & 16.67  & 16.67  & 16.67  & 16.67  \\ \hline
    \end{tabular}}
    \label{tab:SA_acc}
\end{table}

%%%%%%%%%%%%%%%%%%%%%%%%%%%%%%%%%%%

%%%%%%%%%%%%%%%%%%%%%%%%%%%%%%%%%%%%

%%%%%%%%%%%%%%%%%%%%%%%%%%%%%%%%%%
\subsection{Salinas Full Scene Dataset (SFS)}

The Salinas full scene (SFS) dataset was acquired over Salinas Valley California, through an Airborne Visible/Infrared Imaging Spectrometer (AVIRIS) sensor. It comprises $512\times217$ pixels per band and a total of $244$ bands with $3.7$ meter spatial resolution. It consists of vineyard fields, vegetables, and bare soils and contains sixteen classes as shown in Table \ref{tab:SAF_stat}. Few water absorption bands $(108-112, 154-167$ and $224)$ are removed from the dataset before analysis. 

\begin{figure}[!hbt]
    \centering
%     \begin{subfigure}{0.085\textwidth}
% 		\includegraphics[width=0.99\textwidth]{images/SFS_ground_truth.eps}
% 		\centering
% 		\caption{SFS}
% 		\label{Fig.SFS_gt}
% 	\end{subfigure}
    \begin{subfigure}{0.085\textwidth}
		\includegraphics[width=0.99\textwidth]{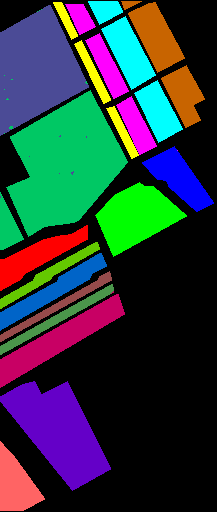}
		\centering
		\caption{$15,9$}
		\label{Fig.2A}
	\end{subfigure}
	\begin{subfigure}{0.085\textwidth}
		\includegraphics[width=0.99\textwidth]{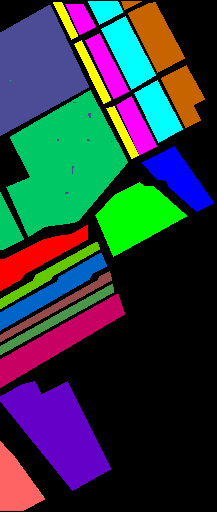}
		\centering
		\caption{$15,11$} 
		\label{Fig.2B}
	\end{subfigure}
	\begin{subfigure}{0.085\textwidth}
		\includegraphics[width=0.99\textwidth]{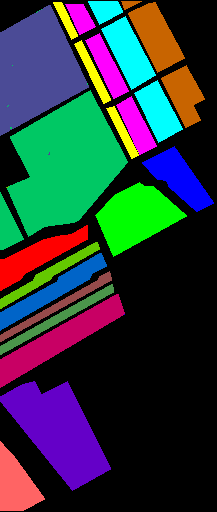}
		\centering
		\caption{$18,9$}
		\label{Fig.2C}
	\end{subfigure}
	\begin{subfigure}{0.085\textwidth}
		\includegraphics[width=0.99\textwidth]{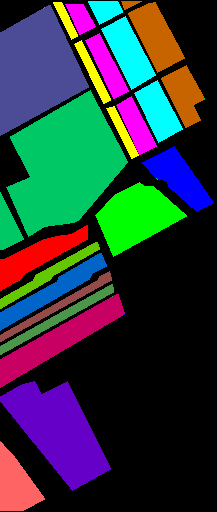}
		\centering
		\caption{$18,11$}
		\label{Fig.2D}
	\end{subfigure}
	\begin{subfigure}{0.085\textwidth}
		\includegraphics[width=0.99\textwidth]{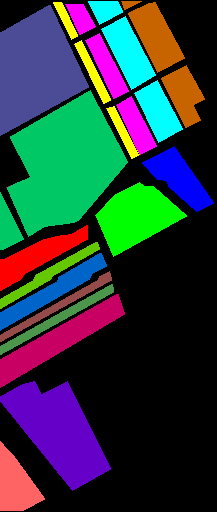}
		\centering
		\caption{$21,9$}
		\label{Fig.2E}
	\end{subfigure}
	\begin{subfigure}{0.085\textwidth}
		\includegraphics[width=0.99\textwidth]{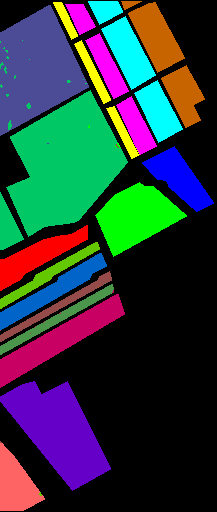}
		\centering
		\caption{$21,11$} 
		\label{Fig.2F}
	\end{subfigure}
	\begin{subfigure}{0.085\textwidth}
		\includegraphics[width=0.99\textwidth]{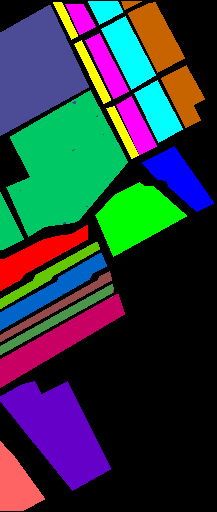}
		\centering
		\caption{$24,9$}
		\label{Fig.2G}
	\end{subfigure}
	\begin{subfigure}{0.085\textwidth}
		\includegraphics[width=0.99\textwidth]{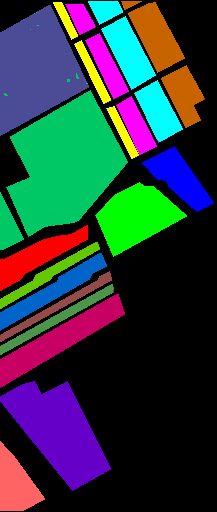}
		\centering
		\caption{$24,11$}
		\label{Fig.2H}
	\end{subfigure}
	\begin{subfigure}{0.085\textwidth}
		\includegraphics[width=0.99\textwidth]{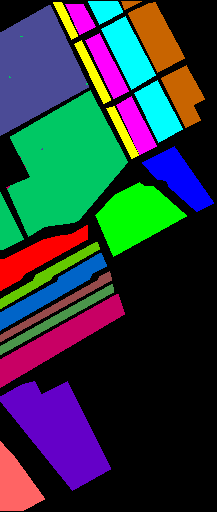}
		\centering
		\caption{$27,9$}
		\label{Fig.2I}
	\end{subfigure}
	\begin{subfigure}{0.085\textwidth}
		\includegraphics[width=0.99\textwidth]{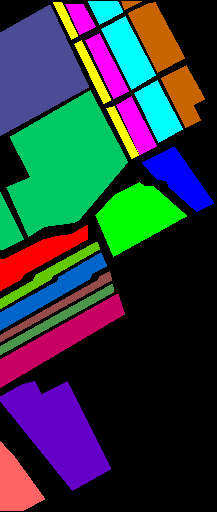}
		\centering
		\caption{$27,11$}
		\label{Fig.2J}
	\end{subfigure}
\caption{Classification results for \textbf{Salinas} for different number of bands ($15, 18, 21, 24, 27$) selected using PCA, with different number of patch sizes ($9\times9$ and $11\times11$).}
\label{Fig.2}
\end{figure}

\begin{figure}[!hbt]
	\begin{subfigure}{0.22\textwidth}
		\includegraphics[width=0.99\textwidth]{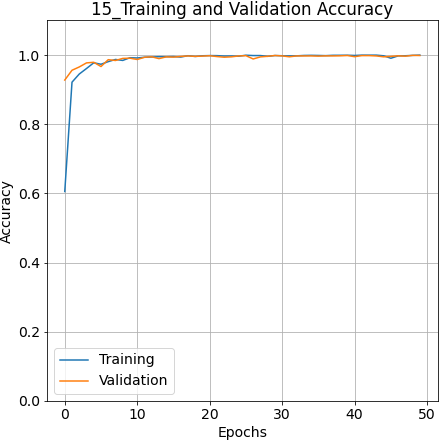}
		\centering
		\caption{Accuracy, $9\times9 \times 15$} 
		\label{Fig.SAF_LCA}
	\end{subfigure}
	\begin{subfigure}{0.22\textwidth}
		\includegraphics[width=0.99\textwidth]{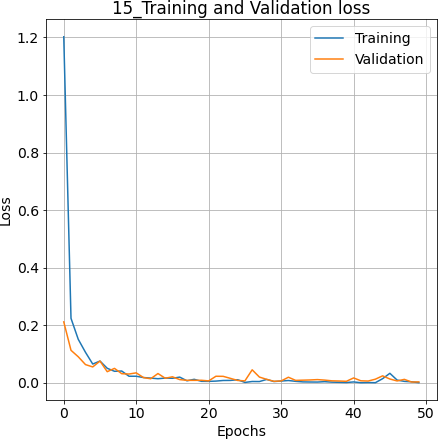}
		\centering
		\caption{Loss, $9\times9 \times 15$}
		\label{Fig.SAF_LCB}
	\end{subfigure}
	
	\begin{subfigure}{0.22\textwidth}
		\includegraphics[width=0.99\textwidth]{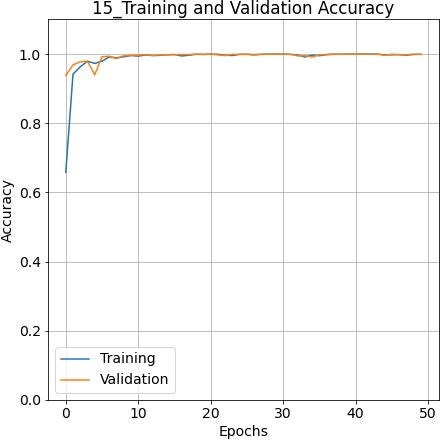}
		\centering
		\caption{Accuracy, $11\times11 \times 15$} 
		\label{Fig.SAF_LCC}
	\end{subfigure}
	\begin{subfigure}{0.22\textwidth}
		\includegraphics[width=0.99\textwidth]{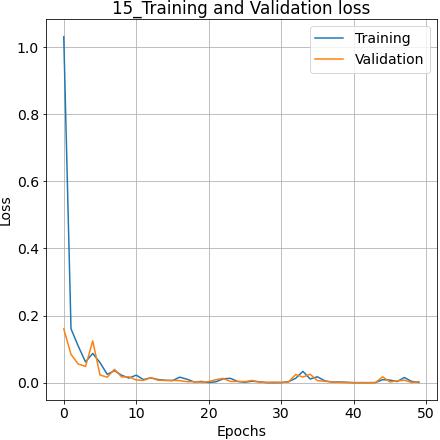}
		\centering
		\caption{Loss, $11\times11 \times 15$} 
		\label{Fig.SAF_LCD}
	\end{subfigure}
\caption{Accuracy and Loss for  Training  and  Validation  sets on  Salinas for $50$ number of epochs with two different spatial dimensions ($9\times 9$ and $11\times 11$) and $15$ number of bands.}
\label{Fig.SAF_LC}
\end{figure}

A detailed experimental results on Salinas full scene dataset is shown in Table \ref{tab:SFS_acc} and Figure \ref{Fig.2}. Moreover, the statistical significance is shown in Table \ref{tab:SAF_stat}. The convergence loss and accuracy of our proposed Hybrid $3D/2D$ CNN for $50$ epochs with two different patch sizes are illustrated in Figure \ref{Fig.SAF_LC}. From these accuracy and loss curves, one can deduce that the accuracy and loss of our proposed model are converged almost around $8$ epoch and $10$ epoch, respectively, for both $9\times 9$ and $11\times 11$ window sizes.

%%%%%%%%%%%%%%%%%%%%%%%%%%%%%%%%%%%%%%%%%%%%%%%%%%%%%
\begin{table}[!hbt]
\centering
\caption{Kappa, Overall and Average accuracy for \textbf{Salinas dataset} with different number of bands ($15, 18, 21, 24, 27$) and different number of patch sizes ($9\times9$ and $11\times11$).}
    \resizebox{\columnwidth}{!}{
    \begin{tabular}{c|cc|cc|cc|cc|cc} \hline
    
    \multirow{2}{*}{\textbf{Method}} &  \multicolumn{2}{c|}{\textbf{15   Bands}} & \multicolumn{2}{c|}{\textbf{18   Bands}} & \multicolumn{2}{c|}{\textbf{21   Bands}} & \multicolumn{2}{c|}{\textbf{24   Bands}} & \multicolumn{2}{c}{\textbf{27   Bands}} \\ \cline{2-11} 
    
     & \textbf{9 $\times$ 9} & \textbf{11 $\times$ 11} & \textbf{9 $\times$ 9} & \textbf{11 $\times$ 11} & \textbf{9 $\times$ 9} & \textbf{11 $\times$ 11} & \textbf{9 $\times$ 9} & \textbf{11 $\times$ 11} & \textbf{9 $\times$ 9} & \textbf{11 $\times$ 11} \\ \hline
    
    \multirow{3}{*}{\textbf{PCA}} & 99.86 & 99.89 & 99.93 & 99.96 & 99.98 & 99.58 & 99.91 & 99.86 & 99.93 & 99.99 \\ \cline{2-11} 
    & 99.87 & 99.90 & 99.93 & 99.97 & 99.99 & 99.62 & 99.92 & 99.88 & 99.93 & 99.99 \\ \cline{2-11} 
    & 99.90 & 99.97 & 99.97 & 99.97 & 99.98 & 99.68 & 99.95 & 99.90 & 99.97 & 99.98 \\ \hline

    \multirow{3}{*}{\textbf{iPCA}} & 97.42 & 21.48 & 91.66 & 42.17 & 97.70 & 1.57  & 93.46 & 84.19 & 84.19 & 91.70 \\ \cline{2-11} 
    & 97.69 & 33.73 & 92.53 & 49.45 & 97.93 & 21.91 & 94.13 & 85.82 & 85.84 & 92.54 \\ \cline{2-11} 
    & 93.35 & 18.66 & 90.33 & 33.55 & 97.61 & 7.37  & 91.46 & 72.51 & 70.23 & 85.30 \\ \hline
    
    \multirow{3}{*}{\textbf{SPCA}} & 95.97 & 33.53 & 96.57 & 88.31 & 99.31 & 83.12 & 96.65 & 69.92 & 97.04 & 81.52 \\ \cline{2-11} 
    & 96.38 & 39.80 & 96.92 & 89.51 & 99.38 & 85.00 & 97.00 & 73.45 & 97.34 & 83.77 \\ \cline{2-11} 
    & 92.13 & 30.96 & 97.25 & 87.04 & 99.32 & 74.04 & 92.26 & 52.74 & 97.46 & 72.27 \\ \hline

    \multirow{3}{*}{\textbf{ICA}}  & 99.08 & 99.56 & 99.17 & 99.93 & 99.07 & 99.77 & 99.72 & 99.75 & 99.83 & 99.93 \\ \cline{2-11} 
    & 99.17 & 99.60 & 99.25 & 99.93 & 99.17 & 99.80 & 99.75 & 99.78 & 99.85 & 99.94 \\ \cline{2-11} 
    & 99.46 & 99.81 & 99.58 & 99.94 & 99.42 & 99.90 & 99.87 & 99.87 & 99.89 & 99.91 \\ \hline

    \multirow{3}{*}{\textbf{SVD}} & 98.02 & 0.00  & 97.14 & 0.00  & 82.59 & 94.16 & 96.22 & 16.91 & 71.79 & 0.00  \\ \cline{2-11} 
    & 98.22 & 20.82 & 97.43 & 13.43 & 84.44 & 94.76 & 96.61 & 32.28 & 74.82 & 20.82 \\ \cline{2-11} 
    & 98.40 & 6.25  & 98.54 & 6.25  & 73.60 & 91.05 & 91.93 & 12.50 & 68.44 & 6.25  \\ \hline

    %\multirow{3}{*}{\textbf{GRP}} & 0.00  & 0.00  & 0.00  & 0.00  & 0.00  & 0.00  & 1.12  & 0.00  & 0.00  & 0.00  \\ \cline{2-11} 
    %& 20.82 & 20.82 & 20.82 & 20.82 & 13.43 & 20.82 & 4.71  & 11.46 & 2.58  & 20.82 \\ \cline{2-11} 
    %& 6.25  & 6.25  & 6.25  & 6.25  & 6.25  & 6.25  & 8.25  & 6.25  & 6.25  & 6.25  \\ \hline
    \end{tabular}}
    \label{tab:SFS_acc}
\end{table}

%%%%%%%%%%%%%%%%%%%%%%%%%%%%%%%%%%%

%%%%%%%%%%%%%%%%%%%%%%%%%%%%%%%%%%%%

%%%%%%%%%%%%%%%%%%%%%%%%%%%%%%%%%%%%%%%
\begin{table*}[!hbt]
\centering
\caption{Class Names, Total Samples, Train, Validation and Test Sample numbers (Tr, Val, Te) along with the Statistical Test (Precision ($P_r$), Recall ($R_c$) \& F1-Score (F1)) results for \textbf{Salinas-A dataset} with PCA as dimensional reduction method over two window sizes ($W_1 = 9\times9$ and $W_2 = 11\times11$).}
\resizebox{\textwidth}{!}{\begin{tabular}{c|c|c|c|c|c|c|c|c|c|c|c|c|c|c|c|c} \hline

\multirow{3}{*}{\textbf{Class}} & \multirow{3}{*}{\textbf{Tr, Val, Te}} & \multicolumn{3}{c|}{\textbf{15 Bands}} & \multicolumn{3}{c|}{\textbf{18 Bands}} & \multicolumn{3}{c|}{\textbf{21 Bands}} & \multicolumn{3}{c|}{\textbf{24 Bands}} & \multicolumn{3}{c}{\textbf{27 Bands}} \\ \cline{3-17} 

& & \textbf{$P_r$} & \textbf{$R_c$} & \textbf{F1} & \textbf{$P_r$} & \textbf{$R_c$} & \textbf{F1} & \textbf{$P_r$} & \textbf{$R_c$} & \textbf{F1} & \textbf{$P_r$} & \textbf{$R_c$} & \textbf{F1} & \textbf{$P_r$} & \textbf{$R_c$} & \textbf{F1} \\ \cline{3-17}

& & \multicolumn{3}{c|}{\textbf{$W_1/W_2$}} & \multicolumn{3}{c|}{\textbf{$W_1/W_2$}} & \multicolumn{3}{c|}{\textbf{$W_1/W_2$}} & \multicolumn{3}{c|}{\textbf{$W_1/W_2$}} & \multicolumn{3}{c}{\textbf{$W_1/W_2$}} \\ \hline
  
    Brocoli green weeds 1 & 98, 97, 196 & 1.00/1.00&1.00/1.00&1.00/1.00 & 1.00/1.00&1.00/1.00&1.00/1.00 & 0.98/1.00&1.00/1.00&0.99/1.00 & 1.00/1.00&1.00/1.00&1.00/1.00 & 1.00/1.00&1.00/1.00&1.00/1.00 \\ \cline{1-17} 
 
    Corn senesced green weeds &	336, 336, 671 & 1.00/1.00&1.00/1.00&1.00/1.00 & 1.00/1.00&1.00/1.00&1.00/1.00 & 1.00/1.00&1.00/1.00&1.00/1.00 & 1.00/1.00&1.00/1.00&1.00/1.00 & 1.00/1.00&1.00/1.00&1.00/1.00 \\ \cline{1-17} 

    Lettuce romaine 4wk &	154, 154, 308 & 1.00/1.00&1.00/1.00&1.00/1.00 & 1.00/1.00&1.00/1.00&1.00/1.00 & 1.00/1.00&0.99/1.00&0.99/1.00 & 1.00/1.00&1.00/1.00&1.00/1.00 & 1.00/1.00&1.00/1.00&1.00/1.00 \\ \cline{1-17} 

    Lettuce romaine 5wk & 381, 382, 762 & 1.00/1.00&1.00/1.00&1.00/1.00 & 1.00/1.00&1.00/1.00&1.00/1.00 & 0.99/1.00&1.00/1.00&1.00/1.00 & 1.00/1.00&1.00/1.00&1.00/1.00 & 1.00/1.00&1.00/1.00&  1.00/1.00 \\ \cline{1-17} 

    Lettuce romaine 6wk &	168, 169, 337 & 1.00/1.00&1.00/1.00&1.00/1.00 & 1.00/1.00&1.00/1.00&1.00/1.00 & 1.00/1.00&0.99/1.00&0.99/1.00 & 1.00/1.00&1.00/1.00&1.00/1.00 & 1.00/1.00&1.00/1.00& 1.00/1.00 \\ \cline{1-17} 

    Lettuce romaine 7wk &	200, 199, 400 & 1.00/1.00&1.00/1.00&1.00/1.00 & 1.00/1.00&1.00/1.00&1.00/1.00 & 1.00/1.00&1.00/1.00&1.00/1.00 & 1.00/1.00&1.00/1.00&1.00/1.00 & 1.00/1.00&1.00/1.00& 1.00/1.00 \\ \hline
\end{tabular}}
\label{tab:SA_stat}
\end{table*}

%%%%%%%%%%%%%%%%%%%%%%%%%%%%%%%%%%%%%%%
\begin{table*}[!hbt]
\centering
\caption{Class Names, Total Samples, Train, Validation and Test Sample  numbers (Tr, Val, Te) along with the Statistical Test (Precision ($P_r$), Recall ($R_c$) \& F1-Score (F1)) results for \textbf{Salinas} with PCA as dimensional reduction method over two window sizes ($W_1 = 9\times9$ and $W_2 = 11\times11$).}
\resizebox{\textwidth}{!}{\begin{tabular}{c|c|c|c|c|c|c|c|c|c|c|c|c|c|c|c|c} \hline

\multirow{3}{*}{\textbf{Class}} & \multirow{3}{*}{\textbf{Tr, Val, Te}} & \multicolumn{3}{c|}{\textbf{15 Bands}} & \multicolumn{3}{c|}{\textbf{18 Bands}} & \multicolumn{3}{c|}{\textbf{21 Bands}} & \multicolumn{3}{c|}{\textbf{24 Bands}} & \multicolumn{3}{c}{\textbf{27 Bands}} \\ \cline{3-17} 

& & \textbf{$P_r$} & \textbf{$R_c$} & \textbf{F1} & \textbf{$P_r$} & \textbf{$R_c$} & \textbf{F1} & \textbf{$P_r$} & \textbf{$R_c$} & \textbf{F1} & \textbf{$P_r$} & \textbf{$R_c$} & \textbf{F1} & \textbf{$P_r$} & \textbf{$R_c$} & \textbf{F1} \\ \cline{3-17}

& & \multicolumn{3}{c|}{\textbf{$W_1/W_2$}} & \multicolumn{3}{c|}{\textbf{$W_1/W_2$}} & \multicolumn{3}{c|}{\textbf{$W_1/W_2$}} & \multicolumn{3}{c|}{\textbf{$W_1/W_2$}} & \multicolumn{3}{c}{\textbf{$W_1/W_2$}} \\ \hline

    Brocoli\_green\_weeds\_1 & 502,	503,	1004 & 1.00/1.00&1.00/1.00&1.00/1.00 & 1.00/1.00&1.00/1.00&1.00/1.00 & 1.00/1.00&1.00/1.00&1.00/1.00 &1 .00/1.00&1.00/1.00& 1.00/1.00 & 1.00/1.00&1.00/1.00&1.00/1.00 \\ \cline{1-17} 

    Brocoli\_green\_weeds\_2 & 932,	931,	1863 & 1.00/1.00&1.00/1.00&1.00/1.00 & 1.00/1.00&1.00/1.00&1.00/1.00 & 1.00/0.99&1.00/1.00&1.00/1.00 & 1.00/1.00&1.00/1.00&1.00/1.00 &1.00/1.00&1.00/1.00& 1.00/1.00 \\ \cline{1-17} 
 
    Fallow &	494	,	494	,	988 & 1.00/1.00   &1.00/1.00    & 1.00/1.00 &   1.00/1.00   &1.00/1.00     &1.00/1.00 & 1.00/1.00   &1.00/1.00     &1.00/1.00 &  1.00/1.00   &1.00/1.00     &1.00/1.00 & 1.00/1.00   &1.00/1.00     &1.00/1.00 \\ \cline{1-17} 
 
    Fallow\_rough\_plow &	348,	349,	697& 1.00/1.00   &1.00/1.00    & 1.00/1.00 &1.00/1.00   &1.00/1.00     &1.00/1.00 &1.00/1.00   &1.00/1.00    & 1.00/1.00 &1.00/1.00   &1.00/1.00    & 1.00/1.00 & 1.00/1.00   &1.00/1.00     &1.00/1.00 \\ \cline{1-17} 
 
    Fallow\_smooth &	669,	670,	1339 & 1.00/1.00   &1.00/1.00&     1.00/1.00 & 1.00/1.00   &1.00/1.00&     1.00/1.00 &1.00/1.00   &1.00/1.00&     1.00/1.00 &1.00/1.00   &1.00/1.00&     1.00/1.00 &1.00/1.00   &1.00/1.00&     1.00/1.00 \\ \cline{1-17} 
 
    Stubble &	990,	989,	1980 &1.00/1.00&   1.00/1.00    & 1.00/1.00 &1.00/1.00  & 1.00/1.00   &  1.00/1.00 &1.00/1.00&   1.00/1.00 &    1.00/1.00 &1.00/1.00&   1.00/1.00 &    1.00/1.00 &1.00/1.00 &  1.00/1.00 &    1.00/1.00 \\ \cline{1-17} 

    Celery &	895,	894,	1790 &1.00/1.00   &1.00/1.00    & 1.00/1.00 &1.00/1.00   &1.00/1.00     &1.00/1.00 &1.00/1.00   &1.00/1.00     &1.00/1.00 &1.00/1.00   &1.00/1.00     &1.00/1.00 &1.00/1.00   &1.00/1.00     &1.00/1.00 \\ \cline{1-17} 
 
    Grapes\_untrained & 2818,	2817,	5636 &1.00/1.00   &1.00/1.00     &1.00/1.00 &1.00/1.00   &1.00/1.00    & 1.00/1.00 &1.00/0.99   &1.00/1.00    & 1.00/0.99 &1.00/1.00   &1.00/1.00     &1.00/1.00 &1.00/1.00   &1.00/1.00    & 1.00/1.00 \\ \cline{1-17} 
 
    Soil\_vinyard\_develop & 1550,	1551,	3102 &1.00/1.00   &1.00/1.00     &1.00/1.00 &1.00/1.00   &1.00/1.00    & 1.00/1.00 &1.00/1.00   &1.00/1.00     &1.00/1.00 &1.00/1.00   &1.00/1.00     &1.00/1.00 &1.00/1.00   &1.00/1.00     &1.00/1.00 \\ \cline{1-17} 
 
    Corn\_senesced\_green\_weeds & 820,	819,	1639 &1.00/1.00   &1.00/1.00     &1.00/1.00 &1.00/1.00   &1.00/1.00     &1.00/1.00 &1.00/1.00   &1.00/1.00    & 1.00/1.00 &1.00/1.00   &1.00/1.00     &1.00/1.00 &1.00/1.00   &1.00/1.00     &1.00/1.00 \\ \cline{1-17} 
 
    Lettuce\_romaine\_4wk & 267,	267,	534 &1.00/1.00   &1.00/1.00    & 1.00/1.00 &0.99/1.00   &1.00/1.00     &1.00/1.00 &1.00/0.99   &1.00/1.00    & 1.00/1.00 &1.00/1.00   &1.00/1.00     &1.00/1.00 &1.00/1.00   &1.00/1.00     &1.00/1.00 \\ \cline{1-17} 
 
    Lettuce\_romaine\_5wk & 482,	482,	963 &1.00/1.00   &1.00/1.00     &1.00/1.00 &1.00/1.00   &1.00/1.00     &1.00/1.00 &1.00/1.00   &1.00/1.00     &1.00/1.00 &1.00/1.00   &1.00/1.00     &1.00/1.00 &1.00/1.00   &1.00/1.00     &1.00/1.00 \\ \cline{1-17} 
 
    Lettuce\_romaine\_6wk &	229,	229,	458 &1.00/1.00   &1.00/1.00     &1.00/1.00 &1.00/1.00   &1.00/1.00     &1.00/1.00 &1.00/0.98   &1.00/1.00     &1.00/0.99 &1.00/1.00   &1.00/1.00     &1.00/1.00 &1.00/1.00   &1.00/1.00     &1.00/1.00 \\ \cline{1-17} 
 
    Lettuce\_romaine\_7wk &	267,	268,	535 &1.00/1.00   &0.99/1.00     &1.00/1.00 &1.00/1.00   &1.00/1.00     &1.00/1.00 &1.00/1.00   &1.00/0.98     &1.00/0.99 &1.00/1.00   &1.00/0.99    &1.00/1.00 &1.00/1.00   &1.00/1.00    &1.00/1.00 \\ \cline{1-17} 

    Vinyard\_untrained & 1817,	1817,	3634 &1.00/0.99   &1.00/1.00     &1.00/1.00 &1.00/1.00   &1.00/1.00     &1.00/1.00 &1.00/1.00   &1.00/0.98     &1.00/0.99 &1.00/1.00   &1.00/0.99    & 1.00/1.00 &1.00/1.00   &1.00/1.00     &1.00/1.00 \\ \cline{1-17} 
 
    Vinyard\_vertical\_trellis & 452,	452,	903 &1.00/1.00   &1.00/1.00     &1.00/1.00 &1.00/1.00   &1.00/1.00     &1.00/1.00 &1.00/1.00   &1.00/1.00     &1.00/1.00 &1.00/1.00   &1.00/1.00     &1.00/1.00 &1.00/1.00   &1.00/1.00     &1.00/1.00 \\ \hline
\end{tabular}}
\label{tab:SAF_stat}
\end{table*}

%%%%%%%%%%%%%%%%%%%%%%%%%%%%%%%%%%
\subsection{Indian Pines Dataset (IP)}

Indian Pines (IP) dataset was collected over northwestern Indiana’s test site, Indian Pines, by AVIRIS sensor and comprises of $145\times145$ pixels and $224$ bands in the wavelength range $0.4-2.5\times10^{-6 }$ meters. This dataset consists of two-thirds of agricultural area and one-third of forest or other naturally evergreen vegetation. A railway line, two dual-lane highways, low-density buildings, and housing and small roads are also part of this dataset. Furthermore, some corps in the early stages of their growth is also present with approximately less than $5\%$ of total coverage. The ground truth is comprised of a total of $16$ classes but they all are not mutually exclusive. The details of ground truth classes are shown in Table \ref{tab:IPD_stat}. The number of spectral bands is reduced to $200$ from $224$ by removing the water absorption bands. 

\begin{figure}[!hbt]
    \centering
%     \begin{subfigure}{0.085\textwidth}
% 		\includegraphics[width=0.99\textwidth]{images/IP_ground_truth.eps}
% 		\centering
% 		\caption{IP}
% 		\label{Fig.IP_gt}
% 	\end{subfigure}
    \begin{subfigure}{0.085\textwidth}
		\includegraphics[width=0.99\textwidth]{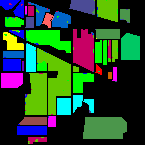}
		\centering
		\caption{$15,9$}
		\label{Fig.3A}
	\end{subfigure}
	\begin{subfigure}{0.085\textwidth}
		\includegraphics[width=0.99\textwidth]{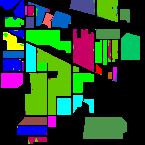}
		\centering
		\caption{$15,11$} 
		\label{Fig.3B}
	\end{subfigure}
	\begin{subfigure}{0.085\textwidth}
		\includegraphics[width=0.99\textwidth]{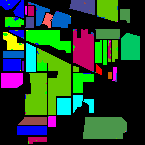}
		\centering
		\caption{$18,9$}
		\label{Fig.3C}
	\end{subfigure}
	\begin{subfigure}{0.085\textwidth}
		\includegraphics[width=0.99\textwidth]{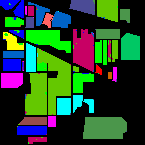}
		\centering
		\caption{$18,11$}
		\label{Fig.3D}
	\end{subfigure}
	\begin{subfigure}{0.085\textwidth}
		\includegraphics[width=0.99\textwidth]{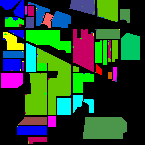}
		\centering
		\caption{$21,9$}
		\label{Fig.3E}
	\end{subfigure}
	\begin{subfigure}{0.085\textwidth}
		\includegraphics[width=0.99\textwidth]{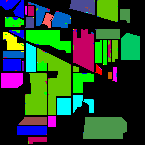}
		\centering
		\caption{$21,11$} 
		\label{Fig.3F}
	\end{subfigure}
	\begin{subfigure}{0.085\textwidth}
		\includegraphics[width=0.99\textwidth]{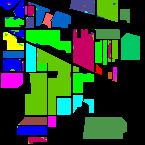}
		\centering
		\caption{$24,9$}
		\label{Fig.3G}
	\end{subfigure}
	\begin{subfigure}{0.085\textwidth}
		\includegraphics[width=0.99\textwidth]{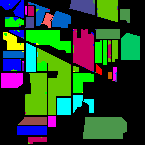}
		\centering
		\caption{$24,11$}
		\label{Fig.3H}
	\end{subfigure}
	\begin{subfigure}{0.085\textwidth}
		\includegraphics[width=0.99\textwidth]{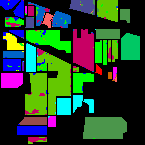}
		\centering
		\caption{$27,9$}
		\label{Fig.3I}
	\end{subfigure}
	\begin{subfigure}{0.085\textwidth}
		\includegraphics[width=0.99\textwidth]{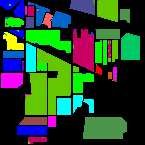}
		\centering
		\caption{$27,11$}
		\label{Fig.3J}
	\end{subfigure}
\caption{Classification results for \textbf{Indian Pines} for different number of bands ($15, 18, 21, 24, 27$) with $9\times9$ and $11\times11$ patch sizes respectively.}
\label{Fig.3}
\end{figure}

\begin{figure}[!hbt]
	\begin{subfigure}{0.22\textwidth}
		\includegraphics[width=0.99\textwidth]{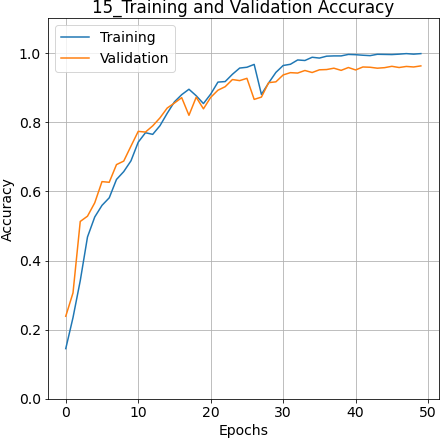}
		\centering
		\caption{Accuracy, $9\times9 \times 15$} 
		\label{Fig.IPCA}
	\end{subfigure}
	\begin{subfigure}{0.22\textwidth}
		\includegraphics[width=0.99\textwidth]{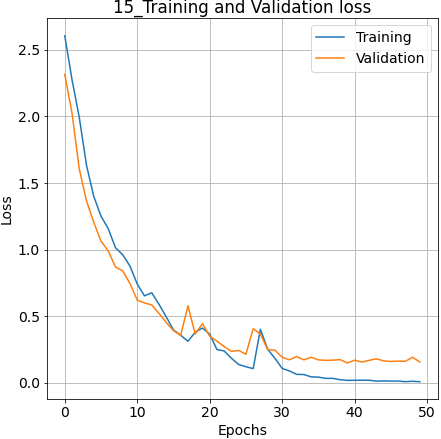}
		\centering
		\caption{Loss, $9\times9 \times 15$}
		\label{Fig.IPCB}
	\end{subfigure}
	
	\begin{subfigure}{0.22\textwidth}
		\includegraphics[width=0.99\textwidth]{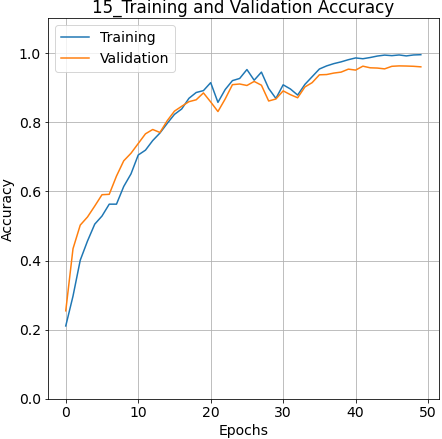}
		\centering
		\caption{Accuracy, $11\times11 \times 15$}
		\label{Fig.IPCC}
	\end{subfigure}
	\begin{subfigure}{0.22\textwidth}
		\includegraphics[width=0.99\textwidth]{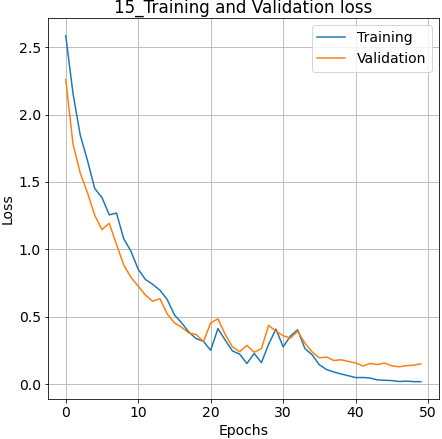}
		\centering
		\caption{Loss, $11\times11 \times 15$}
		\label{Fig.IPCD}
	\end{subfigure}
\caption{Accuracy and Loss for  Training  and  Validation  sets on Indian Pines for $50$ number of epochs with two different spatial dimensions ($9\times 9$ and $11\times 11$) and $15$ number of bands.}
\label{Fig.IPC}
\end{figure}

The convergence loss and accuracy of our proposed Hybrid $3D/2D$ CNN for $50$ epochs with two different patch sizes are illustrated in Figure \ref{Fig.IPC}. From these accuracy and loss curves, one can deduce that our proposed model is converged almost around $35$ epoch for both $9\times 9$ and $11\times 11$ window sizes. A detailed experimental results on Indian Pines dataset is shown in Table \ref{tab:IP_acc} and Figure \ref{Fig.3}. Moreover, the statistical significance is shown in Table \ref{tab:IPD_stat}.

%%%%%%%%%%%%%%%%%%%%%%%%%%%%%%%%%
\begin{table}[!hbt]
\centering
\caption{Kappa, Overall and Average accuracy for \textbf{Indian Pines} with different number of bands ($15, 18, 21, 24, 27$) and different number of patch sizes (i.e., $9\times9$ and $11\times11$). }
    \resizebox{\columnwidth}{!}{
    \begin{tabular}{c|cc|cc|cc|cc|cc} \hline
    
    \multirow{2}{*}{\textbf{Method}} &  \multicolumn{2}{c|}{\textbf{15   Bands}} & \multicolumn{2}{c|}{\textbf{18   Bands}} & \multicolumn{2}{c|}{\textbf{21   Bands}} & \multicolumn{2}{c|}{\textbf{24   Bands}} & \multicolumn{2}{c}{\textbf{27   Bands}} \\ \cline{2-11} 
    
    & \textbf{9 $\times$ 9} & \textbf{11 $\times$ 11} & \textbf{9 $\times$ 9} & \textbf{11 $\times$ 11} & \textbf{9 $\times$ 9} & \textbf{11 $\times$ 11} & \textbf{9 $\times$ 9} & \textbf{11 $\times$ 11} & \textbf{9 $\times$ 9} & \textbf{11 $\times$ 11} \\ \hline

    \multirow{3}{*}{\textbf{PCA}} & 96.75 & 96.84 & 96.75 & 97.00 & 97.55 & 96.95 & 97.53 & 97.24 & 91.54 & 97.00 \\ \cline{2-11} 
    &  97.15 & 97.23 & 97.15 & 97.37 & 97.85 & 97.33 & 97.83 & 97.58 & 92.57 & 97.37 \\ \cline{2-11} 
    &  97.54 & 95.57 & 97.51 & 93.39 & 97.37 & 96.98 & 97.49 & 95.96 & 91.53 & 96.95 \\ \hline
    
    \multirow{3}{*}{\textbf{iPCA}} & 62.30 & 35.54 & 18.81 & 66.48 & 83.29 & 73.33 & 38.91 & 86.50 & 0.00  & 84.72 \\ \cline{2-11} 
    & 66.87 & 47.47 & 36.27 & 70.71 & 85.40 & 76.82 & 49.89 & 88.18 & 23.96 & 86.63 \\ \cline{2-11} 
    & 56.35 & 25.41 & 12.49 & 46.91 & 67.58 & 50.32 & 31.16 & 81.44 & 6.25  & 75.68 \\ \hline

    \multirow{3}{*}{\textbf{SPCA}} & 73.20 & 75.66 & 76.12 & 81.86 & 68.95 & 11.82 & 73.96 & 79.81 & 0.00  & 75.88 \\ \cline{2-11} 
    & 76.72 & 78.87 & 79.08 & 84.16 & 72.88 & 28.45 & 77.23 & 82.44 & 23.96 & 78.79 \\ \cline{2-11} 
    & 65.43 & 56.02 & 64.46 & 74.97 & 59.91 & 12.27 & 65.39 & 60.81 & 6.25  & 53.97 \\ \hline

    \multirow{3}{*}{\textbf{ICA}}  & 68.52 & 71.24 & 65.13 & 71.36 & 79.84 & 84.42 & 91.83 & 90.23 & 79.94 & 90.55 \\ \cline{2-11} 
    & 72.72 & 75.10 & 69.76 & 75.36 & 82.42 & 86.40 & 92.84 & 91.41 & 82.48 & 91.75 \\ \cline{2-11} 
    & 60.79 & 59.92 & 57.74 & 61.50 & 70.80 & 81.28 & 86.59 & 85.93 & 74.69 & 85.49 \\ \hline

    \multirow{3}{*}{\textbf{SVD}}  & 14.77 & 0.00  & 0.00  & 0.00  & 0.00  & 0.00  & 0.00  & 29.94 & 0.00  & 0.00  \\ \cline{2-11} 
    & 30.79 & 23.96 & 23.96 & 23.96 & 23.96 & 23.96 & 23.96 & 43.28 & 23.96 & 23.96 \\ \cline{2-11} 
    & 12.34 & 6.25  & 6.25  & 6.25  & 6.25  & 6.25  & 6.25  & 18.70 & 6.25  & 6.25  \\ \hline
    
    %\multirow{3}{*}{\textbf{GRP}}  & 0.00  & 0.00  & 0.00  & 0.00  & 0.00  & 0.00  & 0.00  & 0.00  & 0.00  & 0.00  \\ \cline{2-11} 
    %& 12.35 & 7.12  & 12.35 & 12.35 & 23.96 & 23.96 & 12.35 & 23.96 & 7.12  & 23.96 \\ \cline{2-11} 
    %& 6.25  & 6.25  & 6.25  & 6.25  & 6.25  & 6.25  & 6.25  & 6.25  & 6.25  & 6.25  \\ \hline
    \end{tabular}}
\label{tab:IP_acc}
\end{table}

%%%%%%%%%%%%%%%%%%%%%%%%%%%%%%%%%%%

%%%%%%%%%%%%%%%%%%%%%%%%%%%%%%%%%%%%

%%%%%%%%%%%%%%%%%%%%%%%%%%%%%%%%%%%%%%%
\begin{table*}[!hbt]
\centering
\caption{Class Names, Total Samples, Train, Validation and Test Sample  numbers (Tr, Val, Te) along with the Statistical Test (Precision ($P_r$), Recall ($R_c$) \& F1-Score (F1)) results for \textbf{Indian Pines dataset} with PCA as dimensional reduction method over two window sizes i.e., $W_1 = 9\times9$ and $W_2 = 11\times11$.}
\resizebox{\textwidth}{!}{\begin{tabular}{c|c|c|c|c|c|c|c|c|c|c|c|c|c|c|c|c} \hline

\multirow{3}{*}{\textbf{Class}} & \multirow{3}{*}{\textbf{Tr, Val, Te}} & \multicolumn{3}{c|}{\textbf{15 Bands}} & \multicolumn{3}{c|}{\textbf{18 Bands}} & \multicolumn{3}{c|}{\textbf{21 Bands}} & \multicolumn{3}{c|}{\textbf{24 Bands}} & \multicolumn{3}{c}{\textbf{27 Bands}} \\ \cline{3-17} 

& & \textbf{$P_r$} & \textbf{$R_c$} & \textbf{F1} & \textbf{$P_r$} & \textbf{$R_c$} & \textbf{F1} & \textbf{$P_r$} & \textbf{$R_c$} & \textbf{F1} & \textbf{$P_r$} & \textbf{$R_c$} & \textbf{F1} & \textbf{$P_r$} & \textbf{$R_c$} & \textbf{F1} \\ \cline{3-17}

& & \multicolumn{3}{c|}{\textbf{$W_1/W_2$}} & \multicolumn{3}{c|}{\textbf{$W_1/W_2$}} & \multicolumn{3}{c|}{\textbf{$W_1/W_2$}} & \multicolumn{3}{c|}{\textbf{$W_1/W_2$}} & \multicolumn{3}{c}{\textbf{$W_1/W_2$}} \\ \hline

    Alfalfa & 11, 12, 23 & 1.00/1.00   &0.96/0.91    & 0.98/0.95 &1.00/1.00   &0.96/0.91     &0.98/0.95 &1.00/1.00  & 0.96/0.87     &0.98/0.93 &1.00/1.00   &1.00/0.96    & 1.00/0.98 &1.00/0.96   &1.00/0.96     &1.00/0.96 \\ \cline{1-17} 

    Corn-notill & 357, 357, 714 & 0.96/0.96   & 0.96/0.98    & 0.96/0.97 & 0.95/0.95   & 0.95/0.96 & 0.95/0.96 &0.98/0.96 & 0.96/0.97 & 0.97/0.96 & 0.99/0.98   & 0.96/0.97 & 0.97/0.97 & 0.94/0.98 & 0.85/0.94 & 0.89/0.96 \\ \cline{1-17} 
 
    Corn-mintill & 208, 207, 415 &0.96/0.97   &0.98/0.98     &0.97/0.97 &0.94/0.95   &0.99/0.98    & 0.96/0.96 &0.95/0.95   &0.99/0.97    & 0.97/0.96 &0.96/0.95   &0.95/0.98     &0.97/0.96 &0.86/0.94   &0.97/0.98     &0.91/0.96 \\ \cline{1-17} 
 
    Corn & 59, 60, 118 &0.99/1.00   &0.95/0.89     &0.97/0.94 &0.97/1.00   &0.89/0.86     &0.93/0.92 &0.94/0.96   &0.93/0.92     &0.94/0.94 &1.00/0.94   &0.92/0.96     &0.96/0.95 &0.96/1.00   &0.84/0.93     &0.90/0.96 \\ \cline{1-17} 

    Grass-pasture & 121, 120, 242 &1.00/0.99   &0.98/0.95     &0.99/0.97 &0.99/0.99   &0.96/0.99     &0.97/0.99 &1.00/0.99   &0.97/0.98     &0.98/0.99 &0.97/0.99   &0.97/0.95     &0.97/0.97 &1.00/0.98   &0.93/0.95     &0.97/0.97 \\ \cline{1-17} 

    Grass-trees & 183, 182, 365 &1.00/0.99   &0.99/1.00     &0.99/1.00 &1.00/0.99   &1.00/1.00     &1.00/1.00 &1.00/0.99   &1.00/1.00     &1.00/1.00 &0.98/0.99   &1.00/1.00    & 0.99/1.00 &0.98/1.00   &0.99/1.00     &0.99/1.00 \\ \cline{1-17} 

    Grass-pasture-mowed & 7, 7, 14 &1.00/0.92   &1.00/0.86     &1.00/0.89 &1.00/1.00   &1.00/1.00    & 1.00/1.00 &0.93/0.93   &1.00/1.00     &0.97/0.97 &0.64/0.93   &1.00/1.00     &0.78/0.97 &0.82/0.74   &1.00/1.00    & 0.90/0.85 \\ \cline{1-17} 

    Hay-windrowed & 120, 119, 239 &1.00/0.99   &1.00/1.00     &1.00/0.99 &1.00/1.00   &1.00/1.00    & 1.00/1.00 &1.00/1.00   &1.00/1.00     &1.00/1.00 &1.00/1.00   &1.00/1.00     &1.00/1.00 &1.00/1.00   &1.00/1.00     &1.00/1.00 \\ \cline{1-17} 

    Oats & 5, 5, 10 &0.91/0.90   &1.00/0.90     &0.95/0.90 &0.83/1.00 &  1.00/0.40 &0.91/0.57 &1.00/1.00   &0.90/1.00     &0.95/1.00 &1.00/1.00 &  0.90/0.70     &0.95/0.82 &0.83/0.90   &0.50/0.90     &0.62/0.90 \\ \cline{1-17} 
 
    Soybean-notill & 243, 243, 486 &0.96/0.95   &0.93/0.94     &0.94/0.95 &0.99/0.98&   0.93/0.93     &0.96/0.95 &0.99/0.95   &0.95/0.94     &0.97/0.95 &0.96/0.99   &0.96/0.93     &0.96/0.96 &0.78/0.97   &0.90/0.94    & 0.83/0.95 \\ \cline{1-17} 

    Soybean-mintill & 614, 613, 1228 &0.97/0.98   &0.98/0.97     &0.97/0.98 &0.98/0.98&   0.97/0.98     &0.98/0.98 &0.98/0.98   &0.99/0.98     &0.98/0.98 &0.98/0.98&   0.99/0.99     &0.98/0.98 &0.94/0.97   &0.91/0.99     &0.93/0.98 \\ \cline{1-17} 

    Soybean-clean & 148, 148, 297 &0.94/0.92   &0.94/0.96     &0.94/0.94 &0.93/0.94 &  0.98/0.98     &0.95/0.96 &0.96/0.95   &0.98/0.94     &0.97/0.94 &0.96/0.95   &0.99/0.97     &0.97/0.96 &0.92/0.95   &0.86/0.97     &0.89/0.96 \\ \cline{1-17} 
 
    Wheat & 51, 52, 102 &1.00/1.00   &0.98/1.00     &0.99/1.00 &0.97/1.00&   1.00/0.98     &0.99/0.99 &1.00/0.96   &0.99/0.99     &1.00/0.98 &1.00/1.00   &0.98/0.99     &0.99/1.00 &1.00/1.00   &0.98/0.98     &0.99/0.99 \\ \cline{1-17} 
 
    Woods & 316, 316, 633 &0.99/0.99   &0.99/0.99     &0.99/0.99 &0.99/0.99   &0.99/1.00    & 0.99/0.99 &0.99/1.00   &0.99/1.00     &0.99/1.00 &1.00/0.99   &1.00/1.00     &1.00/1.00 &0.97/0.99   &1.00/1.00     &0.98/1.00 \\ \cline{1-17} 
 
    Buildings-Grass-Trees-Drives & 96, 97, 193 &0.93/0.97   &0.97/0.96     &0.95/0.97 &0.93/0.97   &0.98/0.98     &0.95/0.97 &0.93/0.99   &0.98/0.95     &0.95/0.97 &1.00/0.92   &0.99/0.98     &0.99/0.95 &0.96/0.95   &0.91/0.98     &0.94/0.96 \\ \cline{1-17} 

    Stone-Steel-Towers & 23, 24, 46 &0.92/0.92   &1.00/1.00     &0.96/0.96 &0.92/0.94   &1.00/1.00     &0.96/0.97 &0.90/0.92   &1.00/1.00     &0.95/0.96 &0.98/0.92   &1.00/1.00    & 0.99/0.96 &0.78/0.96   &1.00/1.00     &0.88/0.98 \\ \hline
\end{tabular}}
\label{tab:IPD_stat}
\end{table*}

%%%%%%%%%%%%%%%%%%%%%%%%%%%%%%%%%%
\subsection{Pavia University Dataset (PU)}

This Pavia University (PU) dataset was gathered over Pavia in northern Italy through a Reflective Optics System Imaging Spectrometer (ROSIS) optical sensor during a flight campaign. It consists of $610\times610$ pixels and $103$ spectral bands with a spatial resolution of $1.3$ meters. Some samples in this dataset provide no information and are removed before analysis. The total ground truth classes are $9$ as shown in Table \ref{tab:PUD_stat}.

\begin{figure}[!hbt]
    \centering
%     \begin{subfigure}{0.085\textwidth}
% 		\includegraphics[width=0.99\textwidth]{images/PU_ground_truth.eps}
% 		\centering
% 		\caption{PU}
% 		\label{Fig.PU_gt}
% 	\end{subfigure}
    \begin{subfigure}{0.085\textwidth}
		\includegraphics[width=0.99\textwidth]{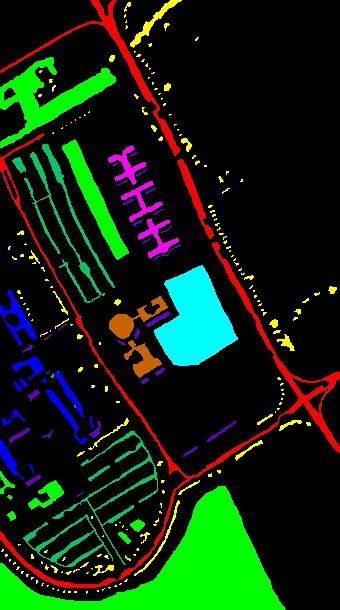}
		\centering
		\caption{$15,9$}
		\label{Fig.5A}
	\end{subfigure}
	\begin{subfigure}{0.085\textwidth}
		\includegraphics[width=0.99\textwidth]{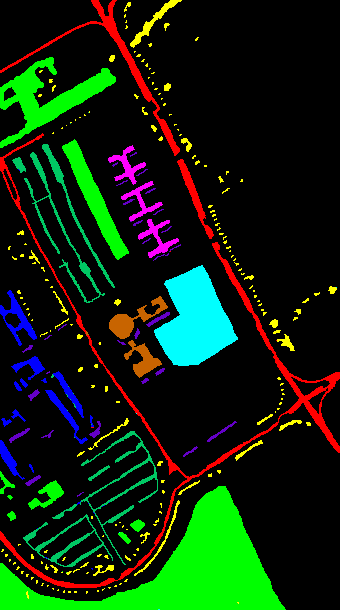}
		\centering
		\caption{$15,11$} 
		\label{Fig.5B}
	\end{subfigure}
	\begin{subfigure}{0.085\textwidth}
		\includegraphics[width=0.99\textwidth]{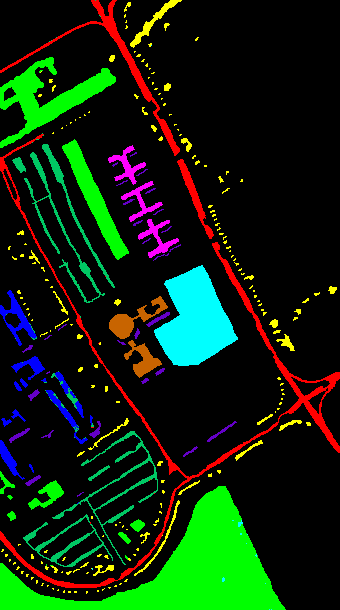}
		\centering
		\caption{$18,9$}
		\label{Fig.5C}
	\end{subfigure}
	\begin{subfigure}{0.085\textwidth}
		\includegraphics[width=0.99\textwidth]{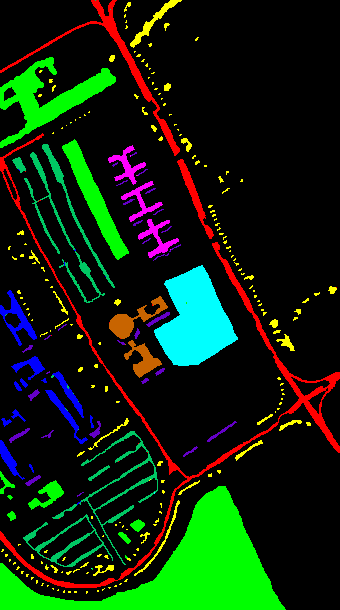}
		\centering
		\caption{$18,11$}
		\label{Fig.5D}
	\end{subfigure}
	\begin{subfigure}{0.085\textwidth}
		\includegraphics[width=0.99\textwidth]{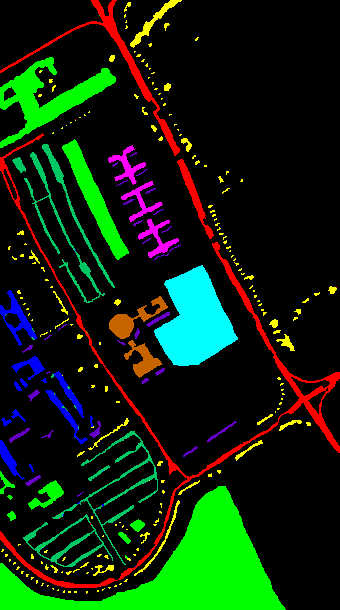}
		\centering
		\caption{$21,9$}
		\label{Fig.5E}
	\end{subfigure}
	\begin{subfigure}{0.085\textwidth}
		\includegraphics[width=0.99\textwidth]{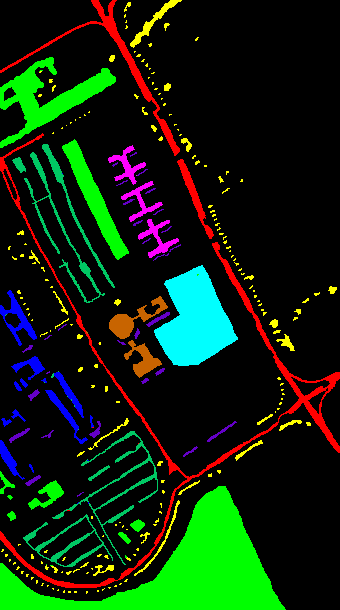}
		\centering
		\caption{$21,11$} 
		\label{Fig.5F}
	\end{subfigure}
	\begin{subfigure}{0.085\textwidth}
		\includegraphics[width=0.99\textwidth]{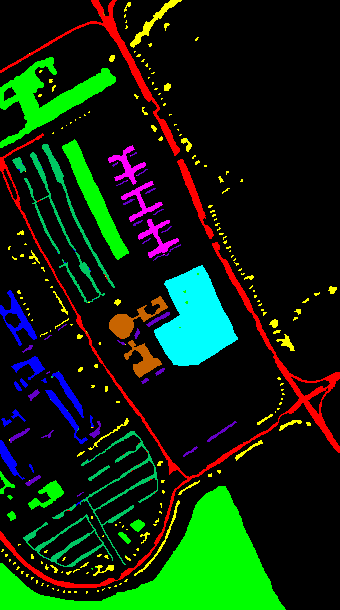}
		\centering
		\caption{$24,9$}
		\label{Fig.5G}
	\end{subfigure}
	\begin{subfigure}{0.085\textwidth}
		\includegraphics[width=0.99\textwidth]{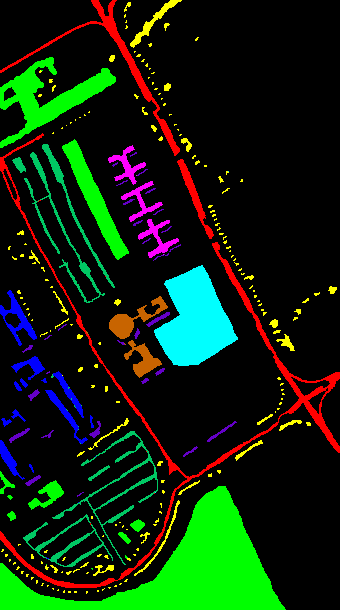}
		\centering
		\caption{$24,11$}
		\label{Fig.5H}
	\end{subfigure}
	\begin{subfigure}{0.085\textwidth}
		\includegraphics[width=0.99\textwidth]{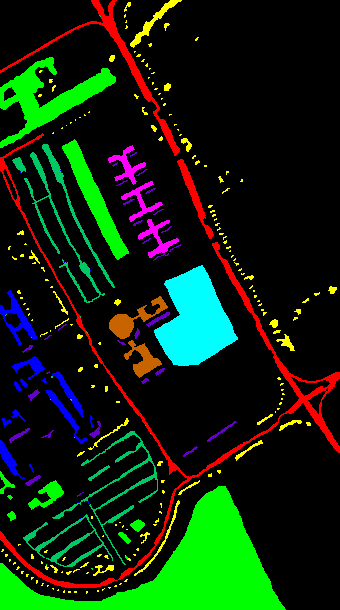}
		\centering
		\caption{$27,9$}
		\label{Fig.5I}
	\end{subfigure}
	\begin{subfigure}{0.085\textwidth}
		\includegraphics[width=0.99\textwidth]{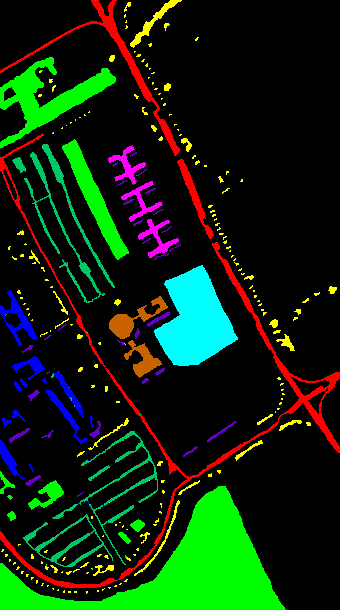}
		\centering
		\caption{$27,11$}
		\label{Fig.5J}
	\end{subfigure}
\caption{Classification results for \textbf{Pavia University dataset} for different number of bands ($15, 18, 21, 24, 27$) selected using PCA, with different number of patch sizes ($9\times9$, and $11\times11$).}
\label{Fig.5}
\end{figure}

\begin{figure}[!hbt]
	\begin{subfigure}{0.22\textwidth}
		\includegraphics[width=0.99\textwidth]{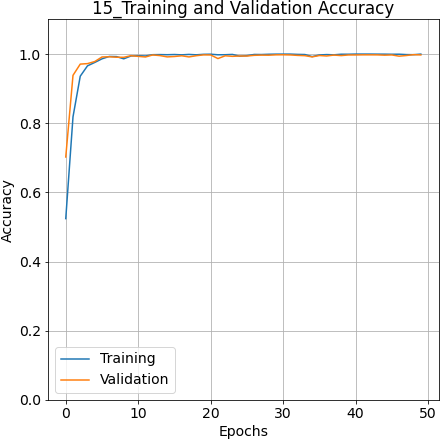}
		\centering
		\caption{Accuracy, $9\times9 \times 15$} 
		\label{Fig.PUA}
	\end{subfigure}
	\begin{subfigure}{0.22\textwidth}
		\includegraphics[width=0.99\textwidth]{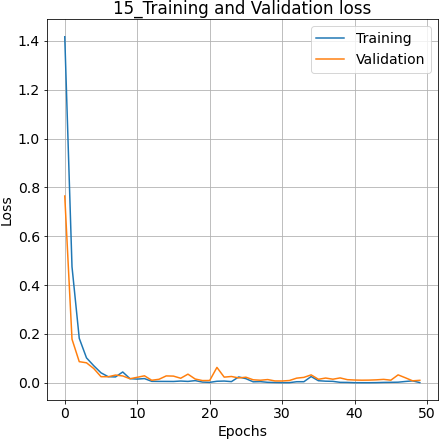}
		\centering
		\caption{Loss, $9\times9 \times 15$}
		\label{Fig.PUB}
	\end{subfigure}
	
	\begin{subfigure}{0.22\textwidth}
		\includegraphics[width=0.99\textwidth]{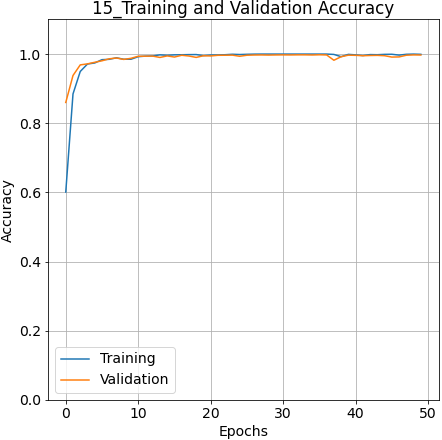}
		\centering
		\caption{Accuracy, $11\times11 \times 15$}
		\label{Fig.PUC}
	\end{subfigure}
	\begin{subfigure}{0.22\textwidth}
		\includegraphics[width=0.99\textwidth]{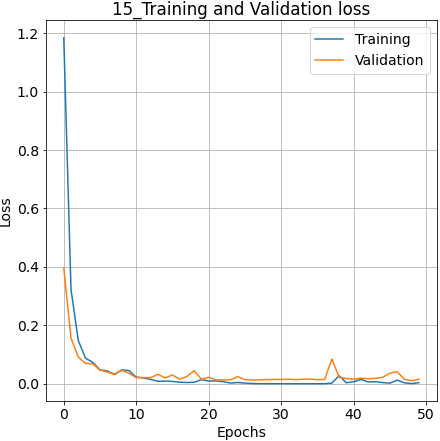}
		\centering
		\caption{Loss, $11\times11 \times 15$}
		\label{Fig.PUD}
	\end{subfigure}
\caption{Accuracy and Loss for  Training  and  Validation  sets on Pavia University for $50$ number of epochs with two different spatial dimensions ($9\times 9$ and $11\times 11$) and $15$ number of bands.}
\label{Fig.PU_LC}
\end{figure}

The convergence loss and accuracy of our proposed Hybrid $3D/2D$ CNN for $50$ epochs with two different patch sizes are illustrated in Figure \ref{Fig.PU_LC}. From these accuracy and loss curves, one can deduce that our proposed model is converged almost around $10$ epoch for both $9\times 9$ and $11\times 11$ window sizes. A detailed experimental result on Pavia University dataset is shown in Table \ref{tab:PU_acc} and Figure \ref{Fig.5}. Moreover, the statistical significance is shown in Table \ref{tab:PUD_stat}.

%%%%%%%%%%%%%%%%%%%%%%%%%%%%%%%%%
\begin{table}[!hbt]
\centering
\caption{Kappa, Overall and Average accuracy for \textbf{Pavia University} with different number of bands ($15, 18, 21, 24, 27$) and different number of patch sizes ($9\times9$ and $11\times11$).}
    \resizebox{\columnwidth}{!}{
    \begin{tabular}{c|cc|cc|cc|cc|cc} \hline
    
    \multirow{2}{*}{\textbf{Method}} &  \multicolumn{2}{c|}{\textbf{15   Bands}} & \multicolumn{2}{c|}{\textbf{18   Bands}} & \multicolumn{2}{c|}{\textbf{21   Bands}} & \multicolumn{2}{c|}{\textbf{24   Bands}} & \multicolumn{2}{c}{\textbf{27   Bands}} \\ \cline{2-11} 
    
    & \textbf{9 $\times$ 9} & \textbf{11 $\times$ 11} & \textbf{9 $\times$ 9} & \textbf{11 $\times$ 11} & \textbf{9 $\times$ 9} & \textbf{11 $\times$ 11} & \textbf{9 $\times$ 9} & \textbf{11 $\times$ 11} & \textbf{9 $\times$ 9} & \textbf{11 $\times$ 11} \\ \hline

    \multirow{3}{*}{\textbf{PCA}} & 56.97 & 99.61 & 58.56 & 99.69 & 56.77 & 99.76 & 59.40 & 99.77 & 64.80 & 99.68 \\ \cline{2-11} 
    & 61.93 & 99.71 & 63.39 & 99.77 & 61.74 & 99.82 & 63.62 & 99.83 & 68.73 & 99.76 \\ \cline{2-11} 
    & 44.55 & 99.54 & 47.32 & 99.55 & 46.56 & 99.69 & 55.41 & 99.64 & 57.55 & 99.59 \\ \hline

    \multirow{3}{*}{\textbf{iPCA}} & 39.35 & 0.00  & 67.85 & 99.11 & 48.34 & 99.40 & 55.18 & 98.77 & 60.45 & 98.93 \\ \cline{2-11} 
    & 46.43 & 43.59 & 71.34 & 99.33 & 54.30 & 99.55 & 60.02 & 99.07 & 64.47 & 99.20 \\ \cline{2-11} 
    & 31.76 & 11.11 & 65.84 & 98.72 & 37.98 & 99.11 & 43.91 & 98.59 & 51.85 & 98.98 \\ \hline

    \multirow{3}{*}{\textbf{SPCA}} & 51.37 & 19.39 & 46.13 & 99.18 & 62.65 & 99.70 & 56.53 & 99.45 & 60.21 & 99.44 \\ \cline{2-11} 
    & 56.45 & 41.50 & 52.53 & 99.38 & 66.50 & 99.78 & 61.28 & 99.59 & 64.74 & 99.57 \\ \cline{2-11} 
    & 42.69 & 15.70 & 34.69 & 98.95 & 55.65 & 99.64 & 46.94 & 99.25 & 47.46 & 99.18 \\ \hline

    \multirow{3}{*}{\textbf{ICA}} & 0.00 & 98.78 & 0.00  & 98.96 & 0.00  & 99.18 & 0.00  & 99.57 & 0.00  & 99.30 \\ \cline{2-11} 
    & 17.81 & 99.08 & 17.81 & 99.21 & 17.81 & 99.38 & 17.81 & 99.67 & 17.81 & 99.47 \\ \cline{2-11} 
    & 7.69 & 98.28 & 7.69  & 98.85 & 7.69  & 99.11 & 7.69  & 99.53 & 7.69  & 99.25 \\ \hline
    
    \multirow{3}{*}{\textbf{SVD}} & 45.29 & 0.00  & 50.72 & 97.13 & 56.33 & 99.08 & 49.49 & 0.00  & 54.06 & 98.57 \\ \cline{2-11} 
    & 52.57 & 43.59 & 57.06 & 97.83 & 61.17 & 99.30 & 56.10 & 43.59 & 59.59 & 98.92 \\ \cline{2-11} 
    & 33.28 & 11.11 & 40.52 & 97.48 & 50.77 & 98.73 & 37.56 & 11.11 & 45.42 & 98.51 \\ \hline

    %\multirow{3}{*}{\textbf{GRP}} & 55.08 & 0.00  & 60.14 & 0.00  & 43.33 & 0.00  & 43.24 & 0.00  & 58.44 & 0.00  \\ \cline{2-11} 
    %& 60.55 & 43.59 & 64.54 & 43.59 & 51.27 & 15.50 & 50.73 & 43.59 & 62.74 & 43.59 \\ \cline{2-11} 
    %& 50.39 & 11.11 & 52.37 & 11.11 & 32.54 & 11.11 & 34.81 & 11.11 & 48.06 & 11.11 \\ \hline
    \end{tabular}}
\label{tab:PU_acc}
\end{table}

%%%%%%%%%%%%%%%%%%%%%%%%%%%%%%%%%%%

%%%%%%%%%%%%%%%%%%%%%%%%%%%%%%%%%%%%

%%%%%%%%%%%%%%%%%%%%%%%%%%%%%%%%%%%%%%%
\begin{table*}[!hbt]
\centering
\caption{Class Names, Total Samples, Train, Validation and Test Sample  numbers (Tr, Val, Te) along with the Statistical Test (Precision ($P_r$), Recall ($R_c$) \& F1-Score (F1)) results for \textbf{Pavia University} with PCA as dimensional reduction method over two window sizes ($W_1 = 9\times9$ and $W_2 = 11\times11$).}
\resizebox{\textwidth}{!}{\begin{tabular}{c|c|c|c|c|c|c|c|c|c|c|c|c|c|c|c|c} \hline

\multirow{3}{*}{\textbf{Class}} & \multirow{3}{*}{\textbf{Tr, Val, Te}} & \multicolumn{3}{c|}{\textbf{15 Bands}} & \multicolumn{3}{c|}{\textbf{18 Bands}} & \multicolumn{3}{c|}{\textbf{21 Bands}} & \multicolumn{3}{c|}{\textbf{24 Bands}} & \multicolumn{3}{c}{\textbf{27 Bands}} \\ \cline{3-17} 

& & \textbf{$P_r$} & \textbf{$R_c$} & \textbf{F1} & \textbf{$P_r$} & \textbf{$R_c$} & \textbf{F1} & \textbf{$P_r$} & \textbf{$R_c$} & \textbf{F1} & \textbf{$P_r$} & \textbf{$R_c$} & \textbf{F1} & \textbf{$P_r$} & \textbf{$R_c$} & \textbf{F1} \\ \cline{3-17}

& & \multicolumn{3}{c|}{\textbf{$W_1/W_2$}} & \multicolumn{3}{c|}{\textbf{$W_1/W_2$}} & \multicolumn{3}{c|}{\textbf{$W_1/W_2$}} & \multicolumn{3}{c|}{\textbf{$W_1/W_2$}} & \multicolumn{3}{c}{\textbf{$W_1/W_2$}} \\ \hline

    Asphalt & 1657, 1658, 3316 &1.00/1.00   &1.00/1.00     &1.00/1.00 &1.00/1.00   &1.00/1.00    & 1.00/1.00 &1.00/1.00   &1.00/1.00     &1.00/1.00 &1.00/1.00   &1.00/1.00     &1.00/1.00 &1.00/1.00   &1.00/1.00    & 1.00/1.00 \\ \cline{1-17} 

    Meadows & 4662, 4663, 9324 &1.00/1.00   &1.00/1.00     &1.00/1.00 &1.00/1.00  & 1.00/1.00     &1.00/1.00 &1.00/1.00  & 1.00/1.00     &1.00/1.00 &1.00/1.00&   1.00/1.00     &1.00/1.00 &1.00/1.00  & 1.00/1.00    & 1.00/1.00 \\ \cline{1-17} 

    Gravel & 525, 525, 1049 &0.99/0.98   &0.99/0.98     &0.99/0.98 &1.00/0.99   &0.88/0.99     &0.93/0.99 &0.99/1.00  & 0.96/0.98    & 0.98/0.99&0.98/1.00   &0.99/0.99     &0.99/0.99 &0.99/0.99   &0.99/0.98     &0.99/0.99 \\ \cline{1-17} 

    Trees & 766, 766, 1532 &1.00/1.00   &0.99/1.00     &1.00/1.00 &1.00/1.00   &0.99/0.99     &0.99/1.00 &1.00/1.00   &1.00/0.99     &1.00/1.00 &1.00/1.00   &1.00/0.99     &1.00/1.00 &1.00/1.00   &1.00/1.00     &1.00/1.00 \\ \cline{1-17} 

    Painted metal sheets & 336, 336, 673 &1.00/1.00   &1.00/1.00    & 1.00/1.00 &1.00/1.00   &1.00/1.00    & 1.00/1.00 &1.00/1.00   &1.00/1.00     &1.00/1.00 &1.00/1.00   &1.00/1.00     &1.00/1.00 &1.00/1.00   &1.00/1.00     &1.00/1.00 \\ \cline{1-17} 
 
    Bare Soil &  1258, 1257, 2514 &1.00/1.00   &1.00/1.00     &1.00/1.00 &0.99/1.00   &1.00/1.00    & 0.99/1.00 &1.00/1.00   &1.00/1.00    & 1.00/1.00 &1.00/1.00   &1.00/1.00     &1.00/1.00 &1.00/1.00   &1.00/1.00     &1.00/1.00 \\ \cline{1-17} 

    Bitumen & 333, 332, 665 &1.00/0.99   &1.00/1.00     &1.00/0.99 &1.00/1.00   &1.00/1.00     &1.00/1.00 &0.99/1.00   &1.00/1.00     &1.00/1.00 &0.99/0.99   &1.00/1.00     &0.99/1.00 &0.99/0.99   &1.00/1.00     &1.00/1.00 \\ \cline{1-17} 

    Self-Blocking Bricks & 921, 920, 1841 &0.99/0.99   &0.99/0.99     &0.99/0.99 &0.93/0.99   &1.00/0.99     &0.96/0.99 &0.98/0.99   &0.99/1.00     &0.99/0.99&1.00/0.99   &0.98/1.00     &0.99/0.99 &0.99/0.99   &0.99/0.99    & 0.99/0.99 \\ \cline{1-17} 

    Shadows & 236, 237, 474 &1.00/0.99   &1.00/1.00    & 1.00/1.00 &1.00/1.00   &1.00/0.99     &1.00/1.00 &1.00/1.00   &1.00/1.00     &1.00/1.00 &1.00/1.00   &0.99/0.99     &1.00/1.00 &1.00/1.00   &1.00/1.00    & 1.00/1.00 \\ \hline
\end{tabular}}
\label{tab:PUD_stat}
\end{table*}

\section{Results Discussion}
\label{Sec:5}

In all the above-discussed experiments, we evaluated the performance of our proposed model for a set of experiments that is initially, we analyzed several dimensionality reductions approaches for hybrid 3D/2D CNN and assessed the performance against a different number of spectral bands ($15, 18, 21, 24$ and $27$) extracted through PCA, iPCA, SPCA, SVD, and ICA methods. Later we examined the effect of input window size on the classification performance of the proposed model by choosing two different patch sizes ($9 \times 9$ and $11 \times 11$).

The experimental results on six benchmark datasets (mentioned in section \ref{Sec:3}) are presented in Tables \ref{tab:SA_acc}, \ref{tab:SFS_acc}, \ref{tab:IP_acc} and \ref{tab:PU_acc} and Figures \ref{Fig.SA_LC} - \ref{Fig.5}. From the results, one can conclude that for all the datasets, the proposed model performed significantly better with PCA as compared to the other well-known dimensionality reduction methods. However, from the experimental results, one can observe that the $\kappa$, OA, AA values remain almost the same with an increasing number of spectral bands extracted through dimensionality reduction techniques.

The classification performance of CNN based HSIC models also relies on the input window size. If the patch size is too small, it decreases the inter-class diversity in samples and if the patch size is set larger then it may take in the pixels from various classes, hence, both cases result in misclassification. We evaluated the proposed framework against two window sizes (i.e. $W_1 = 9 \times 9$ and $W_2 = 11 \times 11$). From the experimental results of Salinas-A, Salinas Full Scene, and Indian Pines dataset, it can be observed that there is a slight improvement in the classification results with increased window size. However, in the case of Pavia University, and the Botswana dataset, one can notice a considerable enhancement in the classification accuracy with $11 \times 11$ window patch as compared to $9 \times 9$.

%%%%%%%%%%%%%%%%%%%%%%%%%%%%%%%%%%
\section{Comparison with State-of-the-art}
\label{Sec:6}

For comparison purposes, the proposed method is compared with various state-of-the-art frameworks published in recent years. From the experimental results presented in Table \ref{tab:my_label} one can interpret that the proposed Hybrid $3D/2D$ CNN has obtained results comparable to the state-of-the-art frameworks and some extent outperformed with respect to the other models. The comparative frameworks used in this work are Multi-scale-3D-CNN \cite{he2017multi}, 3D/2D-CNN \cite{li2017spectral, Hamida18}. All the comparative models are being trained as per the settings mentioned in their respective papers. The experimental results listed in Table \ref{tab:my_label} shows that the proposed Hybrid $3D/2D$ CNN has significantly improved the classification results as compared to the other methods with less number of convolutional layers, number of filters, number of epochs, even a small number of training samples, and mainly, in less computational time.

%%%%%%%%%%%%%%%%%%%%%%%%%%%%%%%%%%%%
\begin{table*}[!hbt]
    \centering
    \caption{Comparative evaluations with State-of-the-art methods while considering $11 \times 11$ Spatial dimensions with even less number of training samples (i.e., $50/50\%$ (train/test) and $50/50\%$ (train/validation)).}
    \resizebox{\textwidth}{!}{
    \begin{tabular}{c|c|c|c|c|c|c|c|c|c|c|c|c} \hline 
        \multirow{2}{*}{\textbf{Dataset}} & \multicolumn{3}{c|}{\textbf{2D-CNN}} & \multicolumn{3}{c|}{\textbf{3D-CNN}} & \multicolumn{3}{c|}{\textbf{MS-3D-CNN}} & \multicolumn{3}{c}{\textbf{Proposed}} \\ \cline{2-13}  
        
        & OA  & AA  & Kappa  & OA  & AA  & Kappa & OA  & AA  & Kappa & OA  & AA  & Kappa  \\ \hline 
        
        Salinas Full & 96.34 & 94.36 & 95.93 & 85.00 & 89.63 & 83.20 & 94.20 & 96.66 & 93.61 & \textbf{99.89} & \textbf{99.90} & \textbf{99.97} \\ \hline 
        
        Indian Pines & 80.27 & 68.32 & 75.26 & 82.62 & 76.51 & 79.25 & 81.39 & 75.22 & 81.20 & \textbf{96.84} & \textbf{97.23} & \textbf{95.57} \\ \hline 
        
        Botswana & 82.54 & 80.23 & 79.62 & 88.26 & 86.15 & 87.52 & 83.93 & 83.22 & 82.56 & \textbf{97.73} & \textbf{97.91} & \textbf{98.20} \\ \hline 

        Pavia University & 96.63 & 94.84 & 95.53 &  96.34 & 97.03 & 94.90 & 95.95 & 97.52 & 93.40 & \textbf{99.61} & \textbf{99.71} & \textbf{99.54} \\ \hline 
    \end{tabular}}
    \label{tab:my_label}
\end{table*}

%%%%%%%%%%%%%%%%%%%%%%%%%%%%%%%%%%
\section{Conclusion}
\label{Sec:7}

Hyperspectral Image Classification (HSIC) is a challenging task due to the spectral mixing effect which induces high intra-class variability and high inter-class similarity in HSI. Nonetheless, 2D CNN is utilized for efficient spatial feature exploitation and several variants of 3D CNN are used for joint spectral-spatial HSIC. However, 3D CNN is a computationally complex model and 2D CNN alone cannot efficiently extract discriminating spectral features. Therefore, to overcome these challenges, this work presented a Hybrid 3D/2D CNN model that provided outstanding HSI classification results on five benchmark datasets in a computationally efficient manner. To summarize, our end-to-end trained Hybrid 3D/2D CNN framework has significantly improved the classification results as compared to the other state-of-the-art methods while being computationally less complex.

%%%%%%%%%%%%%%%%%%%%%%%%%%%%%%%%%%
\section*{Reproducibility}
\label{Sec:8}
The running code is available at Github "https://github.com/mahmad00". 
%%%%%%%%%%%%%%%%%%%%%%%%%%%%%%%%%%%%%%%%%%
{\footnotesize
\bibliographystyle{IEEEtran}
\bibliography{Sam}}
%%%%%%%%%%%%%%%%%%%%%%%%%%%%%%%%%%%%%%%%%%
\end{document}